\documentclass[conference]{IEEEtran}
\usepackage{times}

\usepackage[numbers]{natbib}
\usepackage{multicol}
\usepackage[bookmarks=true]{hyperref}

\usepackage{colortbl}
\usepackage{array}
\usepackage[utf8]{inputenc}
\usepackage{mathtools}
\usepackage{subcaption}
\usepackage{booktabs} % For better table formatting

\usepackage{multirow}
\usepackage{graphicx}
\usepackage{subcaption}

\usepackage[T1]{fontenc}    % use 8-bit T1 fonts
\usepackage{hyperref}       % hyperlinks
\usepackage{url}            % simple URL typesetting
\usepackage{booktabs}       % professional-quality tables
\usepackage{amsfonts}       % blackboard math symbols
\usepackage{nicefrac}       % compact symbols for 1/2, etc.
\usepackage{microtype}      % microtypography
\usepackage{amsmath}
\usepackage{wrapfig}
\usepackage[ruled,vlined,linesnumbered]{algorithm2e} % algorithm2e package
\usepackage{algpseudocode} 
\usepackage{bbm}
\usepackage{pifont}
\usepackage{graphicx}
\usepackage{amssymb}
\usepackage{caption}%注释设置
\usepackage{subcaption}
\usepackage{multirow}
\usepackage{colortbl}
\usepackage[most]{tcolorbox}
\newtcolorbox{response}[1][]{
  colback=gray!5,
  colframe=black,
  fonttitle=\bfseries,
  coltitle=black,
  }
\definecolor{redshade}{HTML}{e6e6e6}
\definecolor{greyshade}{HTML}{FF0000}
\usepackage{tabularx}
\usepackage{minted}

\usepackage{cuted}    % For the 'strip' environment
\usepackage{capt-of}  % To allow captions outside of figure environments

\usepackage{xcolor}         % colors
\usepackage{multirow}
\usepackage{graphicx}

\usepackage{multicol}
\usepackage{colortbl}
\definecolor{mygray}{gray}{.85}
\definecolor{mylight}{RGB}{255, 247, 247}
\definecolor{myhighlight}{RGB}{193,210,240}
\definecolor{newnodepurple}{RGB}{184,170,237}
\definecolor{longedgered}{RGB}{192,0,0}
\definecolor{shortedgegreen}{RGB}{0,176,80}
\definecolor{verylightgray}{RGB}{240,240,240} 
\usepackage{pifont}
\let\labelindent\relax
\usepackage{enumitem}
\begin{document}

% \title{LongNav-R1: Pushing the Limits of Long-Horizon Decision-Making in VLA for Embodied Navigation}

% \title{RSS Rebuttal to [Paper-ID: 394]\vspace{-1cm}}  % **** Enter the paper title and ID here
% \maketitle
\thispagestyle{empty}

\title{LongNav-R1: Horizon-Adaptive Multi-Turn RL for Long-Horizon VLA Navigation}

% \author{Author Names Omitted for Anonymous Review. Paper-ID [394]}
\author{Yue Hu$^{1}$, Avery Xi$^{1}$, Qixin Xiao$^{1}$, Seth Isaacson$^{1}$, 
Henry X. Liu$^{1}$, Ram Vasudevan$^{1}$, Maani Ghaffari$^{1}$ \\
$^1$ University of Michigan, Ann Arbor \\
  \texttt{\{huyu,axi,qxiaocs,sethgi,henryliu,ramv,maanigj\}@umich.edu} 
}

\maketitle

\begin{strip}
    \centering
    \includegraphics[width=\textwidth]{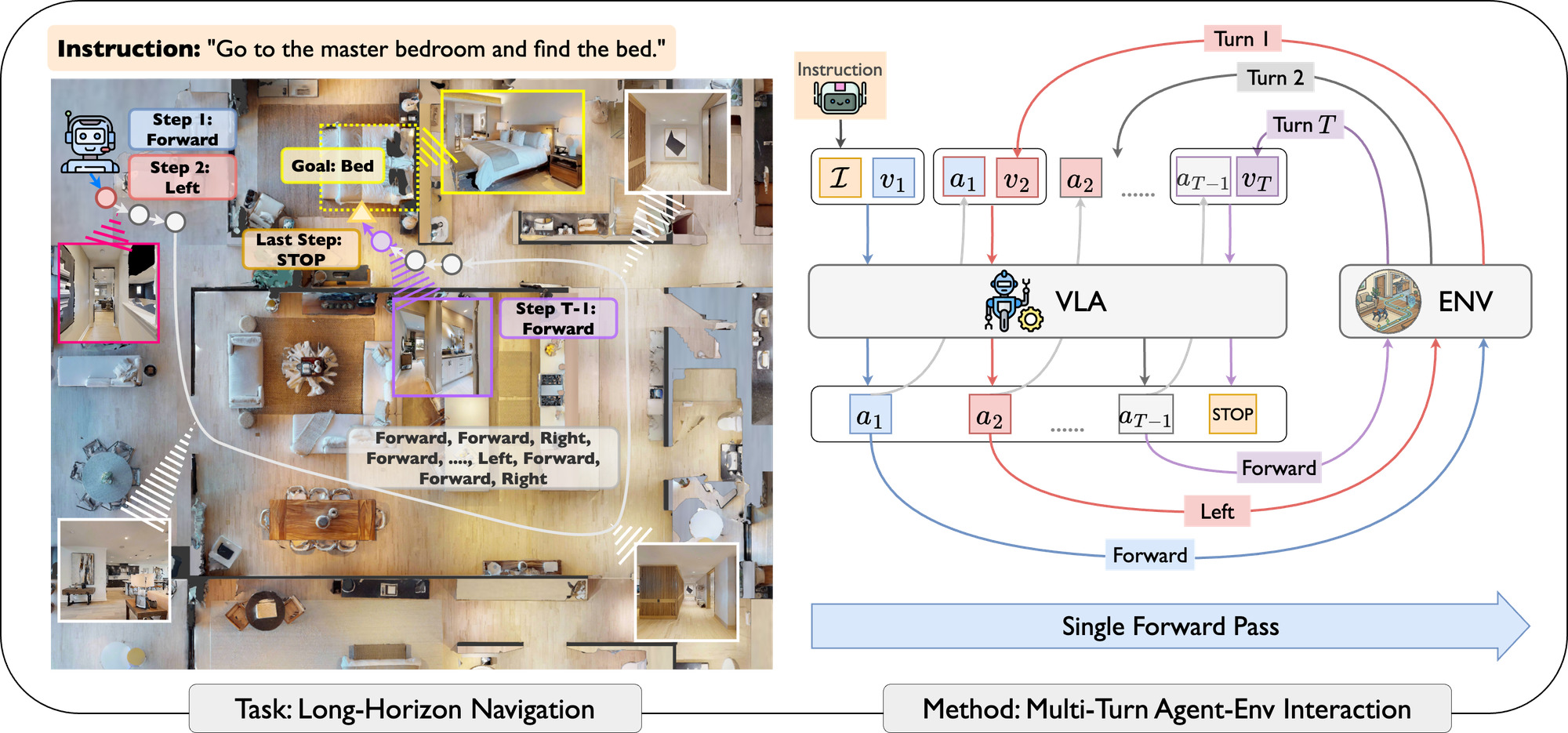}
    \captionof{figure}{LongNav-R1 formulates the navigation process as a multi-turn conversation between the VLA policy and the embodied environment. This end-to-end multi-turn RL framework enables the VLA policy to optimize multi-step decision-making based on cumulative, sequential outcomes.}
    \label{fig:intro}
    \vspace{-4mm}
\end{strip}

% \tableofcontents

\begin{abstract}

This paper develops LongNav-R1, an end-to-end multi-turn reinforcement learning (RL) framework designed to optimize Visual-Language-Action (VLA) models for long-horizon navigation. Unlike existing single-turn paradigm, LongNav-R1 reformulates the navigation decision process as a continuous multi-turn conversation between the VLA policy and the embodied environment. This multi-turn RL framework offers two distinct advantages: i) it enables the agent to reason about the causal effects of historical interactions and sequential future outcomes; and ii) it allows the model to learn directly from online interactions, fostering diverse trajectory generation and avoiding the behavioral rigidity often imposed by human demonstrations. Furthermore, we introduce Horizon-Adaptive Policy Optimization. This mechanism explicitly accounts for varying horizon lengths during advantage estimation, facilitating accurate temporal credit assignment over extended sequences. Consequently, the agent develops diverse navigation behaviors and resists collapse during long-horizon tasks. Experiments on object navigation benchmarks validate the framework's efficacy: With 4,000 rollout trajectories, LongNav-R1 boosts the Qwen3-VL-2B success rate from 64.3\% to 73.0\%. These results demonstrate superior sample efficiency and significantly outperform state-of-the-art methods. The model's generalizability and robustness are further validated by its zero-shot performance in long-horizon real-world navigation settings.
All source code is open-sourced at \url{https://github.com/UMich-CURLY/LongNav-R1}.

\end{abstract}

\IEEEpeerreviewmaketitle

\section{Introduction}

% VLA is promising for Navigation
Navigation is a fundamental capability for intelligent embodied agents, serving as the cornerstone for robots to assist humans in physical environments. 
% To achieve successful navigation, robots must perceive their physical surroundings and move autonomously based on high-level human instructions. 
Historically, navigation systems relied on modular pipelines~\cite{voronav,an2024etpnav,cai2024bridging,zhou2024navgpt,qiao2024opennav} involving separate perception~\cite{liu2024grounding}, mapping~\cite{zhou2025beliefmapnav,yokoyama2024vlfm}, and planning~\cite{SGImagineNav,yin2024sgnav,yin2025unigoal} components. However, recent progress has shifted toward end-to-end Vision-Language-Action (VLA) models~\cite{UniNavid:RSS25,zhang2024navid,wei2025streamvln}. These models leverage large-scale pre-training to enable general semantic navigation, allowing agents to interpret complex visual cues and linguistic commands directly into actions.

% Existing VLA are not deisigned for long-horizon navigation, weakness of existing methods
Despite these advancements, current navigation approaches remain far from achieving human-level performance, particularly in long-horizon tasks. This is because existing state-of-the-art methods~\cite{UniNavid:RSS25,zhang2024navid,liu2025nav,wei2025streamvln} adopt a single-turn imitation learning paradigm, which reduces navigation to a sequence of isolated action predictions based on immediate local context. This formulation introduces two critical deficiencies: first, it lacks causal reasoning by treating steps independently, thereby overlooking the sequential dependencies where early-stage exploration serves as a prerequisite for late-stage, goal-directed efficiency. Second, it leads to behavioral rigidity by strictly imitating expert trajectories instead of optimizing for goal success. Consequently, the agent becomes myopic and brittle, incapable of recovering from errors or adapting to distribution shifts.

\begin{figure}[!t]
    \centering
    \includegraphics[width=1.0\linewidth]{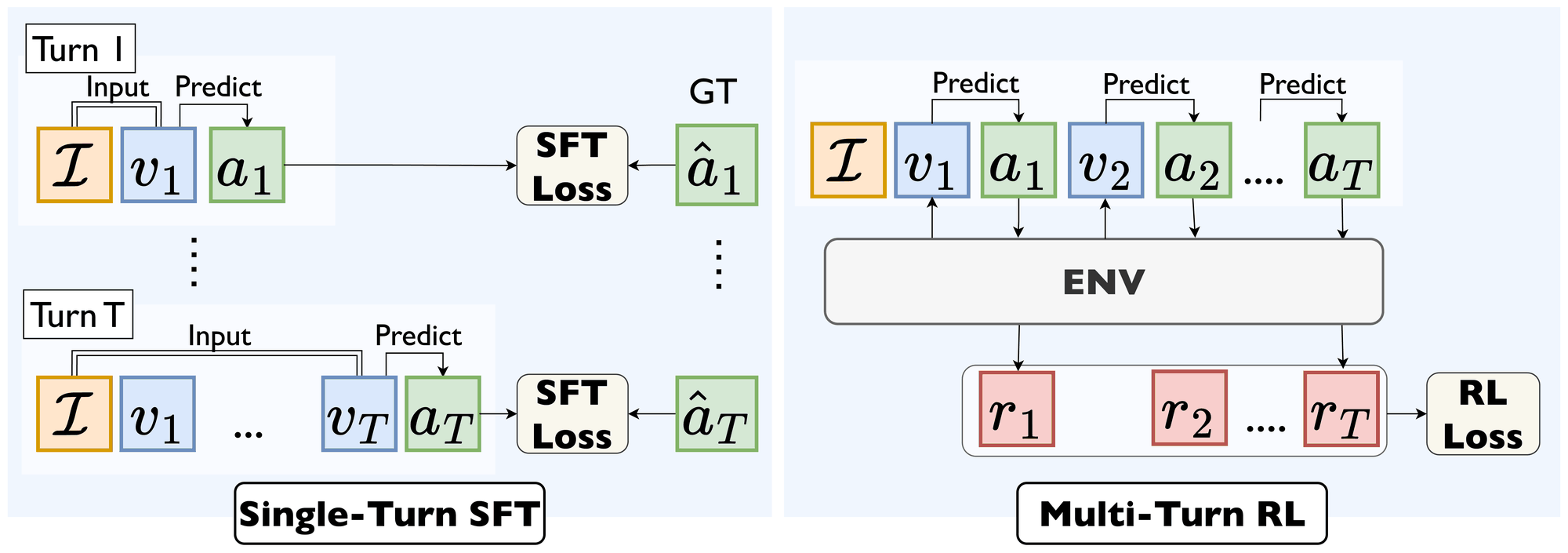}
    \vspace{-5mm}
    \caption{Comparison of single-turn SFT and multi-turn RL.}
    \label{fig:pipeline}
    \vspace{-6mm}
\end{figure}

% We address this issue with a new system
To bridge this gap, this paper proposes LongNav-R1, an end-to-end framework that reformulates navigation as a multi-turn Reinforcement Learning (RL) process. Unlike the single-turn paradigm, LongNav-R1 treats the navigation task as a continuous conversation between the VLA policy and the physical environment. This multi-turn formulation offers two key advantages. First, it provides the model with comprehensive state and objective awareness, allowing it to learn the causal relationship between current actions and distant rewards. Second, by learning directly from online interactions, the agent is encouraged to explore diverse trajectories, thereby overcoming the rigidity of human demonstrations and improving robustness against environmental stochasticity.

While multi-turn RL offers a promising framework for long-horizon VLA navigation, its deployment is bottlenecked by the challenge of temporal credit assignment, as the relative contribution of each turn to the final objective varies significantly over time. While actor-critic methods like PPO~\cite{PPO} manage this via learned value functions, they incur prohibitive computational overhead during long-horizon training. Conversely, efficient critic-free methods from the LLM domain, such as GRPO~\cite{shao2024deepseekmath} and REINFORCE++~\cite{hu2025reinforce++}, are optimized for single-turn tasks with outcome-level rewards and fail to capture the evolving temporal statistics inherent in multi-step robotic decision-making.

% early-stage exploration typically generates high-magnitude rewards, whereas the final target acquisition phase produces sparser, lower-magnitude signals where even minor progress is critical for success
 
% We optimize the system with a now policy gradient
To address this bottleneck, we introduce Horizon-Adaptive Policy Optimization (HAPO), a  framework for critic-free advantage estimation. HAPO eliminates the need for a separate value network by regressing a baseline directly from the local rollout buffer via kernel regression. Drawing inspiration from variance reduction techniques in verifiable RL~\cite{zeng2025shrinking}, we introduce a general formulation that unifies the advantage estimation of critic-free methods with the value function approximation of actor-critic architectures. Under this formulation, existing optimization heuristics such as GRPO and REINFORCE++ emerge as special cases determined by specific kernel choices. By explicitly designing a temporal kernel function, HAPO enables the derived baseline to capture the reward's temporal dynamics of long-horizon navigation, offering the temporal precision of a critic without the associated computational overhead. HAPO enables optimizing VLA models for the varying sequence lengths inherent in long-horizon robotic tasks.

% We validate the performance
To validate the effectiveness of LongNav-R1 and HAPO, we conducted three key evaluations. First, we validated LongNav-R1 across real-world settings and four widely-used simulation benchmarks. Despite being trained on only six object categories in HM3D V1~\cite{hm3dv1}, LongNav-R1 successfully localizes unseen objects in zero-shot real-world scenarios and outperforms previous SOTA methods on HM3D V1~\cite{hm3dv1}, V2~\cite{hm3dv2}, MP3D~\cite{khanna2023hssd} and OVON~\cite{yokoyama2024hm3d}, demonstrating robust generalizability. Second, we conducted ablation studies on core components. Results indicate that multi-turn RL contributes a substantial improvement of 8.7\% on success rate compared to SFT, while HAPO is critical for unlocking long-horizon navigation, boosting success rates from 0\% to 15.4\%. Third, we demonstrated that LongNav-R1 yields highly sample-efficient agents. The success rate of a Qwen3-VL-2B backbone increased from 0.51\% to 73.0\% using only 34k trajectories, outperforming SFT baselines trained on millions of data.

% Furthermore, the model demonstrates robust transfer capabilities when deployed on real-world robots.
 % to 76.0\%, 83.7\%, 63.0\%, and 53.8\%.

% Summary
To summarize, the primary contributions of this paper are:
\begin{enumerate}[leftmargin=*,label=(\roman*)]
% System
\item We propose LongNav-R1, a end-to-end multi-turn reinforcement learning framework that optimizes the VLA policy for multi-step navigation.

% Core technique
\item We introduce HAPO, a critic-free advantage estimation framework that facilitates precise temporal credit assignment, allowing large VLA models to improve multi-step decision-making without the significant computational burden of auxiliary critic networks.

% Experiments
\item We provide a comprehensive experimental validation of LongNav-R1 in real-world and diverse navigation benchmarks, demonstrating that LongNav-R1 significantly outperforms existing methods.
\end{enumerate}
\vspace{-2mm}
\section{Related Works}

\noindent\textbf{Semantic navigation.} Semantic navigation requires agents to navigate to the specific target in unseen environments based on human instructions. There are two representative tasks that involve both visual information and language instructions: Object Goal Navigation~\cite{chaplot2020object,hm3dv1,hm3dv2,Matterport3D} and Vision-and-Language Navigation~\cite{anderson2018r2r,krantz_vlnce_2020,ku2020room}. Early methods~\cite{wijmans2019dd, mousavian2019visual, yang2018visual, majumdar2022zson, maksymets2021thda} largely focused on acquiring task-specific skills via imitation learning~\cite{Habitatweb, ramrakhya2023pirlnav} or RL~\cite{chang2020semantic, gireesh2022object, procthor,zeng2024poliformer}. While these methods can achieve strong performance in trained environments, they often suffer from poor generalization due to domain gaps. Recently, the field has shifted toward leveraging the generalization capabilities of Large-Language Models (LLMs) and Vision-Language Models (VLMs) to improve multi-task navigation. These approaches~\cite{khandelwal2022simple,openai2023gpt4,team2024gemini,anthropic2023claude,Team2025GeminiRB,bi2024deepseek,dubey2024llama,liu2023llava,yu2023l3mvn, zhou2023esc,long2024instructnav,huang2024gamap,yin2024sgnav,UniNavid:RSS25,zhang2024navid,liu2025nav} offer greater flexibility and adaptability in novel environments, but often lack optimized task execution and navigation efficiency, resulting in inferior performance. In contrast, our method trains a VLA model end-to-end with navigation objective, offering both task-aware efficiency and generalization capability.

\noindent\textbf{Large language model for embodied navigation.} 
Large language models have been introduced into robotic navigation due to their generalization capabilities in understanding and planning. Current research follows two primary trajectories. The first utilizes off-the-shelf LLMs and VLMs in a zero-shot fashion. These methods employ LLMs and VLMs as high-level policies to identify landmarks~\cite{yu2024vln,yu2023l3mvn} and select frontiers~\cite{long2024instructnav,long2024instructnav,huang2024gamap,yin2024sgnav}. The second trajectory~\cite{zhang2024navid, UniNavid:RSS25, wei2025streamvln} involves fine-tuning VLMs as end-to-end VLA models that directly generate actions via Supervised Fine-Tuning (SFT). However, SFT-based methods are often bottlenecked by the need for exhaustive expert demonstrations and are susceptible to the domain drift inherent in behavior cloning. To overcome these limitations, we train our VLA model using end-to-end RL, which facilitates the discovery of diverse navigation skills and enhances final navigation performance.

\noindent\textbf{Reinforcement learning for large language model.} Reinforcement learning has become a crucial post-training technique for large language models, facilitating alignment with human preferences~\cite{RLHF} and significantly bolstering reasoning capabilities~\cite{bi2024deepseek}. While these advancements have primarily been applied to web search~\cite{wei2025webagent, xi2025agentgym} and tool-use~\cite{wang2025reinforcement, xue2025simpletir,wang2025ragen}, adapting these benefits to embodied navigation remains an open challenge. A pioneering effort, Nav-R1~\cite{liu2025nav}, adopts the DeepSeek-R1 paradigm~\cite{guo2025deepseek} by generating navigation action sequences in a single turn and training the VLA model via GRPO~\cite{shao2024deepseekmath}. However, embodied navigation is inherently a multi-turn decision-making process involving hundreds of steps; a single-turn formulation fails to capture its sequential and long-horizon nature. To address this, we propose an end-to-end, multi-turn RL framework for VLA models. Furthermore, rather than directly adopting standard advantage estimations from GRPO~\cite{shao2024deepseekmath}, REINFORCE++~\cite{hu2025reinforce++}, or RLOO~\cite{RLOO}, we introduce horizon-adaptive advantage estimation. This technique stabilizes interaction-adaptive optimization and improves long-horizon decision-making performance in embodied navigation.
\vspace{-4mm}
\section{Problem Formulation}

\begin{figure*}[!th]
    \centering
    \includegraphics[width=\linewidth]{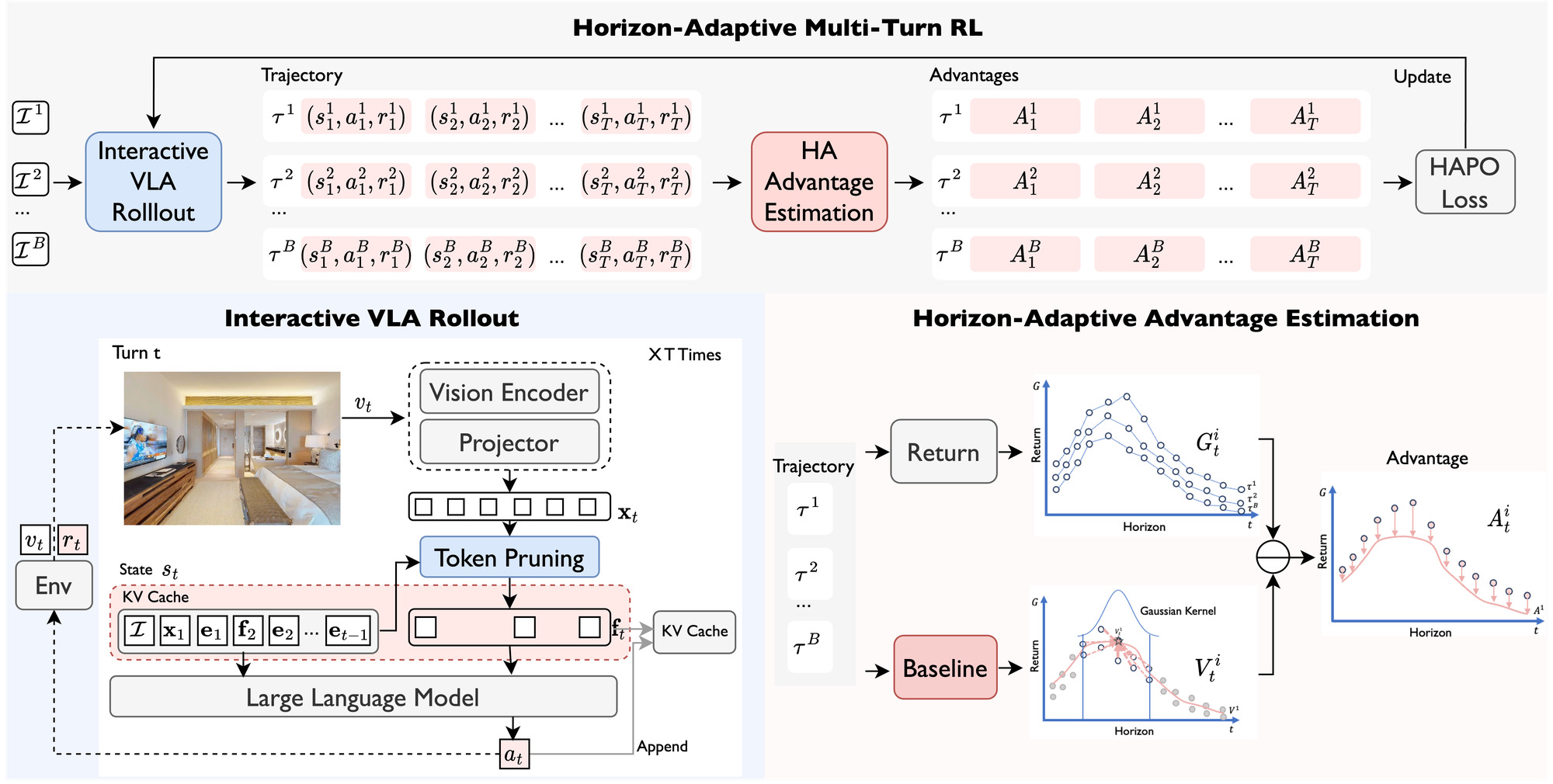}
    \vspace{-5mm}
    \caption{The overall pipeline of LongNav-R1. The framework optimizes the VLA policy through a three-stage iterative process: i) collecting long-horizon trajectories via multi-turn interactive rollout; ii) computing action-level advantages using the proposed horizon-adaptive estimator; and iii) updating the VLA model via the aggregated optimization objective. }
    \label{fig:method}
    \vspace{-5mm}
\end{figure*}

This section establishes the mathematical formulation for robotic navigation using a VLA policy. We first introduce the navigation objective and frame the multi-step navigation decision process as a multi-turn conversation. Then we define a multi-turn RL formulation tailored for VLAs. Building on this foundation, we introduce critic-free horizon-adaptive policy optimization to optimize VLA for long-horizon objective.

\vspace{-2mm}
\subsection{Navigation task objective}

We address the task of general robotic navigation in unknown environments. In this setting, an agent is provided with a open instruction $\mathcal{I}$, describing either a target object or a sequence of steps, and iteratively perceives and interacts with the environment to locate the goal. At each time step $t$, the agent receives an observation $v_t$ and constructs a state $s_t$. Action selection is governed by a parameterized policy $\pi_{\theta}$ conditioned on the current state $s_t$:
\begin{equation}
a_t \sim \pi_{\theta}(\cdot \mid s_t).
\end{equation}
An episode terminates when the agent invokes a `stop' action. Success is defined as reaching the target's vicinity within a fixed step budget. Our objective is to optimize $\pi_{\theta}$ to maximize both navigation efficiency and success rate.

\subsection{Multi-turn RL formulation of VLA policy}

To solve the general navigation tasks, we utilize recent advances in large foundation models to optimize a VLA policy under the navigation objective. Given that navigation is a long-horizon process often spanning hundreds of steps, we frame the task as a multi-turn conversation between the VLA policy and its embodied environment. We model actions auto-regressively, conditioned on the global instruction $\mathcal{I}$, history $\{v_{1:t-1}, a_{1:t-1}\}$, and current observation $v_t$:
\begin{equation}
a_t \sim \pi_{\theta}(\cdot \mid \mathcal{I}, \{v_{1:t-1},a_{1:t-1}\},v_t).
\end{equation}
This multi-turn RL framework offers two advantages. First, it enables KV cache reuse, significantly enhancing computational efficiency during training and inference. The history $\{v_{1:t-1},a_{1:t-1}\}$ is saved in cache and only need to compute new observation. This optimization is critical for long-horizon navigation, which involves hundreds of steps and generates hundreds of thousands of visual tokens. Second, the sequential nature of our formulation allows the model to explicitly account for the causal dependencies of prior interactions. By attributing the final trajectory outcome to individual actions, we improve both learning efficiency and performance on long-horizon tasks.

\noindent\textbf{Comparison with existing VLA policies.} In contrast to UniNavid~\cite{UniNavid:RSS25} and Nav-R1~\cite{liu2025nav}, which adopt a single-turn paradigm, our multi-turn paradigm explicitly models the sequence effects. These prior works predict actions and optimize at the step-level, failing to capture the sequential nature of navigation. Furthermore, while StreamVLN~\cite{wei2025streamvln} utilizes a multi-turn structure, it relies on SFT. Consequently, it suffers from the severe distribution shift issues inherent to behavior cloning and lacks a mechanism to directly optimize for the final, long-term success of the navigation task.

\vspace{-2mm}
\subsection{Horizon-adaptive policy optimization}\label{form:hapo}

To optimize VLA policies for long-horizon navigation, we propose Horizon-Adaptive Policy Optimization (HAPO). While recent advancements such as REINFORCE++~\cite{hu2025reinforce++} and GRPO~\cite{shao2024deepseekmath} have demonstrated efficacy in LLM training, their applicability is constrained by a reliance on single-turn paradigms and sparse, outcome-level rewards. By treating all time steps uniformly, these frameworks fail to capture the temporal dynamics of robotic navigation. In reality, moments like critical turns are far more determinative of success than routine straight-line movements. HAPO addresses this temporal credit assignment challenge by employing a kernel-based advantage estimator that uses step-level dense rewards, allowing the model to explicitly weigh contributions across varying horizons. HAPO has two phases: i) formulating advantage estimation as a generalized non-parametric value estimation framework via kernel regression, and ii) introducing a temporal kernel design to enable horizon-adaptive advantage estimation.

\noindent\textbf{General kernel-based advantage estimation.} Following the theoretical framework established in~\cite{zeng2025shrinking}, we model critic-free baselines as non-parametric value function approximations. Unlike parametric critics that approximate values over the global state space, non-parametric critic-free methods perform local regression based strictly on the current rollout buffer. The key idea is to formalize value estimation as a kernel regression problem over the rollout buffer, which induces a generalized framework for advantage estimation, capable of adapting to diverse task objectives through specialized kernel designs. 

Specifically, let $\mathcal{B}=\{\tau^i\}_{i=1}^B$ denote a buffer of complete rollout trajectories $\tau^i=\{(s_t^i,a_t^i,r_t^i)\}_{t=1}^T$. For each action $a_t^i$ at timestep $t$ in trajectory ${\tau^i}$, the advantage $A_t^i$ is defined as
\begin{small}
\begin{equation}
A_t^i = G_t^i - V_{\mathcal{B}}^K(s_t^i),  ~~ G_t^i = \sum_{t'=t}^{|\tau^{i}|} \gamma^{t'-t} r_{t}  \label{eq:advantage}
\end{equation}
\end{small}
where $G_t^i$ is the return with discount factor $\gamma$, and $V_{\mathcal{B}}^K(\cdot)$ denotes a kernel-based baseline. Specifically, the baseline is computed via leave-one-out kernel regression:
\begin{small}
\begin{equation}
V_{\mathcal{B}}^K(s_t^i)=
\frac{
\displaystyle \sum_{j \neq i}^{B} \sum_{t'=1}^{T}
K\big(f(s_t^i), f(s_{t'}^j)\big)G_{t'}^j
}{
\displaystyle \sum_{j \neq i}^{B} \sum_{t'=1}^{T}
K\big(f(s_t^i), f(s_{t'}^j)\big)
}.
\end{equation}
\end{small}
where $f(\cdot)$ is a feature mapping and $K(\cdot,\cdot)$ is a similarity kernel defined in the feature space. The leave-one-out construction $j\neq i$ excludes samples from the current trajectory to avoid self-bias and ensure unbiased estimation. This formulation subsumes a broad class of critic-free advantage estimators through appropriate choices of the kernel $K$ and feature representation $f$, enabling task-dependent value estimation without learning a parametric critic. For instance, in scenarios governed by temporal dynamics, a time-aware kernel can be employed to capture sequential dependencies; conversely, in environments where geometric configuration is paramount, a spatial-aware kernel is more appropriate for modeling structural regularities. Explicitly integrating these state features yields significantly more accurate value estimates.

\noindent\textbf{Temporal-aware kernel design.} To explicitly account for the sequential structure of navigation, HAPO introduces a time-aware kernel. We account for the temporal dynamics by explicitly defining the discrete time step as the feature, $f(s_i^t) = t$, and employ a Gaussian kernel $K(x, x') = \exp\Big(-\frac{|x - x'|^2}{2\sigma^2}\Big)$ with bandwidth $\sigma$, to provide temporal smoothing for the value estimates. 
% Then the value estimation is,
% \begin{equation}
% \small
% V_{\mathcal{B}}^K(s_t^i)=
% \frac{
% \displaystyle \sum_{j \neq i}^{B} \sum_{t'=1}^{T}
% K\big(f(t), f(t')\big)G_{t'}^j
% }{
% \displaystyle \sum_{j \neq i}^{B} \sum_{t'=1}^{T}
% K\big(f(t), f(t')\big)
% }.\label{eq:value}
% \end{equation}
By applying this formulation, HAPO regresses a baseline that is strictly specific to the current horizon of the episode. This enables the advantage estimator to adapt to the shifting reward distributions inherent in navigation, effectively distinguishing between the high-variance exploration phase (early horizon) and the high-precision target acquisition phase (late horizon).

\noindent\textbf{Comparison with REINFORCE++.} HAPO provides a general advantage estimation framework while existing methods such as REINFORCE++ emerges as special cases. Specifically, REINFORCE++ corresponds to the use of a constant kernel $K(\cdot, \cdot) = 1$ with features defined only at the final outcome state. Under this setting, the value estimate reduces to the mean final reward of the rollout buffer: $V_{\mathcal{B}}^K(s_T^i) = \frac{1}{B}\sum_{j=1}^{B} G_T^j$. This implies that the entire batch shares a uniform baseline, effectively assigning the same value to all actions across all time steps. It introduces significant bias by disregarding the varying step-specific contributions to the final return, a critical oversight in long-horizon navigation tasks.

\section{LongNav-R1: Multi-Turn Navigation Framework}

This section provides a comprehensive overview of our proposed pipeline, illustrated in Fig.~\ref{fig:method}. We begin by detailing the interactive VLA rollout within embodied environments in section~\ref{sub:rollout}, including state encoding, token pruning, and action prediction. Subsequently, we describe our horizon-adaptive advantage estimation in section~\ref{sub:reward}, which processes rollout trajectories to provide dense advantages for long-horizon navigation tasks. Finally, we describe our training strategy in section~\ref{sub:training}, which enables scalable and efficient VLA optimization through SFT warm-up phase followed by multi-turn RL training.

\subsection{Interactive VLA rollout}\label{sub:rollout}

\noindent\textbf{State encoding.} State encoding transforms the multi-modal trajectory history into a sequence of tokens to provide global context for action prediction. Specifically, at each time step $t$, we process the observation-action pair in two stages: i) visual and action encoding. The current visual observation $v_t$ is processed by a vision encoder $\mathbf{E}$ to extract latent features (visual tokens) $\mathbf{x}_t$. Simultaneously, the previous action $a_{t-1}$ is embedded into a textual token $\mathbf{e}_t$ via the tokenizer $\mathbf{D}(\cdot)$:
\begin{equation}
\mathbf{x}_t = \mathbf{E}(v_t), ~~ \mathbf{e}_{t-1} = \mathbf{D}(a_{t-1}).    
\end{equation}

\noindent\textbf{Online token pruning.} We filter redundant visual tokens before updating the state. Since navigation tasks are inherently long-horizon, often involving hundreds of steps, a single trajectory can generate hundreds of thousands of visual tokens. This massive volume creates significant computational overhead. To enhance representation efficiency, we introduce online token pruning at each rollout step. The key idea is to retain only informative visual tokens, defined as those exhibiting low feature similarity to the historical context. 
% Note that, by leveraging the multi-turn paradigm, we utilize the KV cache to represent this historical context, thereby significantly reducing the computation overhead.

Let $\{\mathbf{k}_{1:t-1}, \mathbf{v}_{1:t-1}\}$ be the cached keys and values from the historical trajectory. We compute a binary selection mask $\mathbf{m}_t \in \{0,1\}^{M}$, where an element is set to $1$ only if the token's maximum similarity to the history falls below a predefined threshold $\delta$ (indicating novel information). The retained sparse visual tokens $\mathbf{f}_t$ are computed as:
\begin{equation}
\mathbf{f}_t = \mathbf{m}_t \odot \mathbf{x}_t, ~~ \text{where } \mathbf{m}_t = \mathbb{I}(\max(\mathbf{x}_t \cdot \mathbf{k}_{1:t-1}^\top) < \delta).
\end{equation}
Here, $M$ denotes the number of visual tokens in the current observation, and $\mathbb{I}(\cdot)$ is the indicator function. Only the non-zero (informative) tokens are preserved. Consequently, the sparse state $s_t$ at time step $t$ is updated to include the history cache, the new sparse visual tokens, and the action token:
\begin{equation}
s_t = \{\mathbf{k}_{1:t-1}, \mathbf{v}_{1:t-1}, \mathbf{e}_{t-1}, \mathbf{f}_t\}.
\end{equation}
Note that at the initial step, with the KV cache and action history empty, the model's state is derived from the task instruction $\mathcal{I}$ and the first observation. Subsequently, the state is maintained via the KV cache, which encodes the full history of actions and observations.

\noindent\textbf{Action prediction.} Given the encoded state representation, the agent predicts the optimal next action to advance toward the goal. By encapsulating the full history of observations and actions, the state allows the agent to reason about the causal effects of past decisions and maintain global context, crucial for preventing redundant exploration and enabling targeted search. In the VLA framework, action generation is cast as next-token prediction. At each timestep $t$, the policy outputs a categorical distribution over the vocabulary, from which the action token is sampled:
\begin{equation}
a_t \sim \pi_{\theta}(\cdot|s_t), \quad \text{where } \pi_{\theta}(\cdot|s_t) \in \Delta^{|D|-1}.
\end{equation}
Here, $\pi_{\theta}$ denotes the probability distribution derived from the VLA's softmax-normalized logits, and $|D|$ represents the vocabulary size. While this work focuses on a discrete action space where each command corresponds to a single token, the framework is readily extensible to continuous control via action quantization or tokenization strategies.

Upon predicting action $a_t$, the agent executes it within the embodied environment, prompting a state transition, a reward $r_t$ and a new observation $v_{t+1}$. This sequential rollout iterates until a termination condition, either a stop action or the maximum horizon $T$ is met, yielding the complete trajectory $\tau=\{(s_1,a_1,r_1), (s_2,a_2,r_2), \dots, (s_T,a_T,r_T)\}$. This loop allows the agent to continuously update its state and refine its decision-making to satisfy the navigation instruction.

\subsection{Horizon-adaptive advantage estimation}\label{sub:reward}

To optimize VLA policies within long-horizon, multi-turn settings, we implement HAPO solution proposed in sec.~\ref{form:hapo} for navigation task. The key idea of HAPO is the regression of a temporal-aware baseline via kernel regression over the rollout buffer. This enables accurate advantage estimation that accounts for temporal dynamics. Given full trajectories in the rollout buffer, we compute advantages for every timestep to provide dense process feedback for the VLA policy learning with three main steps.

\noindent\textbf{Return estimation.} Given a full trajectory $\tau^{i}\in\mathcal{B}$, we compute the empirical returns at $t$-th step $G_t^i = \sum_{t'=t}^{|\tau^{i}|} \gamma^{t'-t} r_{t}$ using a predefined discount factor $\gamma$. For the navigation task, we define the step-level reward $r_t$ based on the geodesic distance to the goal, providing a dense proxy for incremental progress. We calibrate the discount factor $\gamma$ to modulate the effective temporal horizon based on environmental semantic density. In indoor settings, high semantic density induces complex, high-frequency temporal dynamics; we therefore restrict the horizon to mitigate variance and maintain a high signal-to-noise ratio. Conversely, sparse outdoor environments exhibit low-frequency temporal dynamics, allowing a larger $\gamma$ to capture long-range dependencies without signal degradation.

% We set the discount factor $\gamma = 0.95$, which targets an effective temporal horizon of approximately 60 steps. 

\noindent\textbf{Kernel-based baseline.} We regress a non-parametric baseline by applying a temporal kernel function across the rollout buffer. To mitigate the high variance inherent in long-horizon tasks, we utilize a critic-free baseline based on Gaussian smoothing. By grouping returns in proximal timestamps, HAPO constructs a stable baseline at each horizon scale. The baseline $V_t^i$ for state $s_t^i$ is calculated as \begin{equation}
    V_t^i = \frac{\sum_{j\neq i}^{B}\sum_{t'=1}^{|\tau_j|} \exp\Big(-\frac{|t - t'|^2}{2\sigma^2}\Big) G_{t'}^j}{\sum_{j\neq i}^{B}\sum_{t'=1}^{|\tau_j|} \exp\Big(-\frac{|t - t'|^2}{2\sigma^2}\Big)}
\end{equation}
where $\mathcal{B}$ represents the rollout buffer and $\sigma$ is the bandwidth parameter controlling the temporal smoothing window. The kernel bandwidth is calibrated to the environment's temporal dynamics. Note that, while standard long-horizon tasks suffer from high variance due to noisy trajectories, our non-parametric kernel regression effectively reduces this variance with temporal smoothing window.

% A narrower smoothing window is employed for high-frequency indoor settings, whereas a wider window is utilized for low-frequency outdoor navigation.
% without introducing the optimization bias inherent in parametric critics.

\noindent\textbf{Dense advantage estimation.} The advantage at each timestep is computed as the residual between the return and the temporal baseline: $A_t^i = G_t^i - V_t^i$. By calculating dense advantage signals at each timestamp, HAPO provides the VLA policy with informative process signals. This allows the model to learn complex navigation dynamics and maintain a precise correspondence between visual-linguistic inputs and sequential actions. This approach significantly enhances sampling efficiency and training stability compared to standard outcome-reward critic-free methods that lack temporal awareness.

\subsection{Training strategy}\label{sub:training}

% Given that RL performance is highly sensitive to implementation specifics,  we adopt a two-stage optimization strategy to ensure stability and sample efficiency. 
We utilize an off-the-shelf VLM~\cite{yang2025qwen3} as our backbone. Since generic VLMs lack embodied navigation priors, we first employ a warm-start phase using imitation learning (IL) on human demonstrations~\cite{Habitatweb}. Subsequently, we transition the agent to an Online RL phase, optimizing the policy via our HAPO. 

% This two-stage approach allows the agent to first acquire semantic navigation primitives via supervision, and then refine these behaviors through extensive exploration, thereby mitigating the covariate shift inherent in pure imitation learning.

\noindent\textbf{Balanced IL Warm-up.} In this phase, we utilize teacher forcing to bootstrap the VLA policy with fundamental navigation behaviors. We process the full trajectory sequence in a single forward pass, computing gradients exclusively on the generated action tokens. The objective is to minimize the standard negative log-likelihood:
\begin{equation}
\mathcal{L}_{\rm IL}(\theta) = -\sum_{i=1}^{B}\sum_{t=1}^{T} \log \pi_\theta(a_t^i \mid \mathcal{I}^{i}),
\end{equation}
where $\mathcal{I}^{i}$ represents the $i$-th instruction in the each batch with size $B$. This supervised pre-training instills essential search strategies, such as panoramic scanning and collision avoidance. However, naive IL induces a strong length bias, particularly affecting the `stop' action; agents tend to terminate episodes near the average length of training trajectories regardless of goal proximity.

% To alleviate this, we employ length-stratified sampling: we balance the data distribution by uniformly sampling trajectories across varying lengths (0 to 400 steps). This ensures the policy learns to terminate based on visual evidence of the target rather than temporal priors.

\noindent\textbf{Online multi-turn RL training.} Following the warm-up, the agent enters the multi-turn RL stage. Here, the objective is to maximize the expected cumulative reward over the full horizon:
\begin{equation}
\resizebox{\linewidth}{!}{$\begin{aligned}
\mathcal{J}_{\text{HAPO}}(\theta) = \mathbb{E}_{\{\tau^i\}_{i=1}^B} \Bigg[ \frac{1}{B} \sum_{i=1}^{B} \frac{1}{|\tau^i|} \sum_{t=1}^{|\tau^i|} \bigg( & \min \left( \rho^i_t A^{\text{norm}, i}_t, \text{clip}(\rho^i_t, 1-\epsilon_{l}, 1+\epsilon_{h}) A^{\text{norm},i}_t \right) \\
& - \beta D_{\text{KL}}\left(\pi_\theta(a^i_t|\mathcal{I}^i) \parallel \pi_{\text{ref}}(a^i_t|\mathcal{I}^i)\right) \bigg) \Bigg].
\end{aligned}$}
\end{equation}
where $A_t^{\text{norm}, i}$ denotes the normalized advantage, computed via global batch normalization to enhance training stability. Following REINFORCE++~\cite{hu2025reinforce++}, the transformation is defined as: $A_t^{\text{norm}, i} = \frac{A_t^i - \text{mean}(A \mid A \in \mathcal{B})}{\text{std}(A \mid A \in \mathcal{B})} + \epsilon$, where $\epsilon$ is a small constant for numerical stability.
And $\rho^i_t = \pi_\theta(a^i_t | \mathcal{I}^i) / \pi_{\text{ref}}(a^i_t | \mathcal{I}^i)$ denotes the importance sampling ratio. The hyperparameters $\epsilon_l$, $\epsilon_h$ and $\beta$ serve to constrain the gradient range and policy update magnitude, respectively, ensuring stable optimization by preventing large architectural shifts during learning. The reference model $\pi_{\rm ref}$ and KL loss $D_{\text{KL}}$ are used for regularization to make the training more stable. We also use the entropy masking trick in ~\cite{cui2025entropy}. Note that, comparing to REINFORCE++, our advantage estimator moves beyond simple batch averages to a time-aware baseline, enabling more precise credit assignment. This allows the policy to better manage stochastic dynamics and rectify errors accumulated during the imitation phase, ensuring robustness over long execution horizons.

\section{Experiments}

\subsection{Experimental setup}

We conduct experiments to evaluate LongNav-R1 on three specific aspects: i) How does LongNav-R1 perform compared to existing state-of-the-arts (SOTA)? ii) Does LongNav-R1 improve long-horizon decision-making capacity for VLA? iii) Is the key design of our method effective? 

% We evaluate in both \textbf{real-world} and simulated environments, encompassing indoor and outdoor scenarios that requires long-horizon embodied navigation. To ensure a comprehensive assessment, we test in our created real-world scenarios and outdoor simulation dataset, OutNav, and two widely used public simulation benchmarks,

\noindent\textbf{Benchmarks.} To evaluate our general-purpose navigation method, we conduct extensive experiments on object-goal and open-vocabulary tasks across four benchmarks: HM3D V1~\cite{hm3dv1}, V2~\cite{hm3dv2}, MP3D~\cite{Matterport3D}, and HM3D-OVON~\cite{yokoyama2024hm3d}. Unlike previous baselines that often prioritize either efficiency or open-vocabulary capabilities on limited subsets, we provide a comprehensive comparison across diverse baselines.

\noindent\textbf{Evaluation metrics.} We report two metrics including \emph{success rate (SR)} and \emph{success rate weighted by path length (SPL)}~\cite{spl}. SR is the core metric of object-goal navigation task, representing the success rate of navigation episodes. SPL measures the ability of the agent to find the optimal path. If success, $\textrm{SPL}=\frac{\textrm{optimal path length}}{\textrm{path length}}$, otherwise $\textrm{SPL}=0$. Higher is better for both metrics.

\noindent\textbf{Implementation details.} For real-world, an Orbbec Femto Bolt sensor is mounted approximately 1m off the ground. To mitigate the sim-to-real gap, the real-world RGB camera is calibrated to approximate the intrinsics of the simulated camera used during training. For simulation, we follow the standard settings~\cite{hm3dv1}. The camera of the agent is $0.88m$ above the ground. The camera outputs $640\times480$ RGB images. The success distance threshold is set as 1m, and the step budget is 500. The discrete action set includes MOVE FORWARD (0.25m), TURN LEFT/RIGHT (30°), and STOP. Ablations are conducted on randomly sampled 200 episodes. For the VLA policy training, we start with a Qwen-3-VL-2B~\cite{yang2025qwen3} as base.

\noindent\textbf{Computation resource.} Both SFT and RL stages were trained on 4 A100 GPUs (72 GPUh per stage). For consistency, efficiency metrics were benchmarked on a single A100. In real-world experiments, an A100 server handles inference, with ROS managing observation and action communication.

% The SFT stage processed 30k trajectories in 18 hours, while RL stage required 18 hours for 4k episodes; this lower throughput stems from online rollout latency and training-rollout orchestration. 
% , and images are undistorted and transformed via a homography 
% \textbf{Baselines.} To perform comprehensive evaluation, we compare our method with three mainstream state-of-the-art methods, spanning training-required methods, zero-shot methods, and visual language model-based methods (both trainable and zero-shot). Specifically, for training-required methods, we compare with PoLiformer~\cite{zeng2024poliformer}, which uses a transformer and RL-based training. For zero-shot methods, we consider CLIP-based VLFM~\cite{yokoyama2024vlfm} and the LLM-prompting-based InstructNav~\cite{long2024instructnav}. Lastly, for the latest visual language model with a trainable approach, we include StreamVLN (16 frames)~\cite{wei2025streamvln}.

\subsection{Comparison with SOTAs}

\begin{table}[t]
% \vspace{-4.5mm}
\centering
\setlength{\tabcolsep}{3pt} 
\caption{\textbf{Comparison on object goal navigation.} LongNav-R1 outperforms previous SOTAs across different metrics on object goal navigation benchmark HM3D V1~\cite{hm3dv1}, V2~\cite{hm3dv2} and MP3D~\cite{Matterport3D}.}
% \renewcommand{\arraystretch}{0.3}
% \vspace{-2mm}
\resizebox{\linewidth}{!}{
\begin{tabular}{lccccccccc}
\toprule
        \multirow{2}*{\textbf{Method}} & \multicolumn{3}{c}{\textbf{Observations}} &
        \multicolumn{2}{c}{\textbf{HM3D V1}}& \multicolumn{2}{c}{\textbf{HM3D V2}} & \multicolumn{2}{c}{\textbf{MP3D}} \\
        \cmidrule(lr){2-4} \cmidrule(lr){5-6} \cmidrule(lr){7-8} \cmidrule(lr){9-10}
        & Odom. & Depth & RGB & SR$\uparrow$ & SPL$\uparrow$ & SR$\uparrow$ & SPL$\uparrow$ & SR$\uparrow$ & SPL$\uparrow$\\
        \midrule
CoWs~\cite{gadre2023cows}  & $\checkmark$ & $\checkmark$ & $\checkmark$  & - & -  & - & - \\
DD-PPO~\cite{wijmans2019dd} & $\checkmark$ & $\checkmark$ & $\checkmark$ & 27.9 & 14.2 & - & - &- &- \\
ESC~\cite{zhou2023esc}    & $\checkmark$ & $\checkmark$ & $\checkmark$ & 39.2 & 22.3 &- &- & 28.7 & 14.2 \\
Habitat-Web~\cite{Habitatweb} & $\checkmark$ & $\checkmark$ & $\checkmark$ & 41.5 & 16.0 &- &-  & 31.6 & 8.5 \\
VoroNav~\cite{voronav}  & $\checkmark$ & $\checkmark$ & $\checkmark$ & 42.0 & 26.0 & - & - &- &-  \\
L3MVN~\cite{yu2023l3mvn}    & $\checkmark$ & $\checkmark$ & $\checkmark$ & 50.4 & 23.1  &- &- & 34.9 & 14.5\\
OpenFMNav~\cite{kuang2024openfmnav} & $\checkmark$ & $\checkmark$ & $\checkmark$  & 52.5 & 24.1  &- &- & 37.2 & 15.7 \\
VLFM~\cite{yokoyama2024vlfm} & $\checkmark$ & $\checkmark$ & $\checkmark$ & 52.5 & 30.4 &- &- & 36.4 & 17.5  \\
GAMap~\cite{huang2024gamap} & $\checkmark$ & $\checkmark$ & $\checkmark$  & 53.1 & 26.0 & - & - & - & - \\
SG-Nav~\cite{yin2024sgnav}  & $\checkmark$ & $\checkmark$ & $\checkmark$ & 54.0 & 24.9 &- &-  & 40.2 & 16.0 \\
UniGoal~\cite{yin2025unigoal}  & $\checkmark$ & $\checkmark$ & $\checkmark$  & 54.5 & 25.1 &- &-  & 41.0 & 16.4 \\
InstructNav~\cite{long2024instructnav} & $\checkmark$ & $\checkmark$ & $\checkmark$  & 58.0 & 20.9 &- &-  & - & - \\
SGM~\cite{zhang2024imagine} & $\checkmark$ & $\checkmark$ & $\checkmark$ &- &- & 60.2 & 30.8  & 37.7 & 14.7 \\ 
BeliefMapNav~\cite{zhou2025beliefmapnav} & $\checkmark$ & $\checkmark$ & $\checkmark$ & 61.4 & 30.6  &- &- & 37.3 & 17.6  \\ 
SGImagineNav~\cite{SGImagineNav} & $\checkmark$ & $\checkmark$ & $\checkmark$ & 65.4 & 30.0 & - & - & - & - \\ 
\midrule
ImagineNav~\cite{zhao2024imaginenav} & $\checkmark$ & & $\checkmark$ & 53.0 & 23.8 & - & - & - & - \\
ProcTHOR~\cite{procthor}  & $\checkmark$ &  & $\checkmark$ & 54.4 & 31.8 & - & - &- &- \\
OVRL~\cite{yadav2023offline}  & $\checkmark$ &  & $\checkmark$ & 62.0 & 26.8 &- &- & 28.6 & 7.4 \\
OVRL-v2 ~\cite{yadav2023ovrl} & $\checkmark$ &  & $\checkmark$ & 64.7 & 28.1 & - & - &- &- \\
PIRLNav-IL~\cite{ramrakhya2023pirlnav} & $\checkmark$ &  & $\checkmark$ & 64.1 & 27.1 & 52.0  & 20.6 &- &- \\
PIRLNav-RL~\cite{ramrakhya2023pirlnav} & $\checkmark$ &  & $\checkmark$ & 70.4 & 34.1 & 61.9 & 27.9 &- &- \\
\midrule
ZSON~\cite{majumdar2022zson}  & &  & $\checkmark$ & 25.5 & 12.6  &- &- & 15.3 & 4.8\\
PixNav~\cite{cai2024bridging} & &  & $\checkmark$  & 37.9 & 20.5 & - & - &- &-\\ 
PSL~\cite{sun2024prioritized}  & &  & $\checkmark$ & 42.4 & 19.2 &- &- & 18.9 & 6.4 \\
UniNavid~\cite{UniNavid:RSS25}& &  & $\checkmark$ & 73.7 & 37.1 & - & - &- &- \\ \midrule
\rowcolor{mylight} \textbf{LongNav-R1} & &  & $\checkmark$ & \textbf{76.0} & \textbf{44.3}  & \textbf{83.7} & \textbf{43.4} & \textbf{63.0} & \textbf{30.2}\\
\bottomrule
\end{tabular}}
% note: mp3d at 1394 episodes.
\vspace{-3mm}
\label{tab:main_hm3d}
\end{table}

% , whereas \textbf{LongNav-R1}$^*$ denotes the model
\begin{table}[t]
\centering
\caption{\textbf{Comparison on open-vocabulary object goal navigation.} Zero-shot LongNav-R1 outperforms previous SOTAs across different metrics on OVON~\cite{yokoyama2024hm3d}. ZS denotes zero-shot. \textbf{LongNav-R1} is trained only on HM3D without fine-tuning on the OVON. Val-Seen-Syn is the Val-Seen-Synonyms split.}
\resizebox{\linewidth}{!}{
\begin{tabular}{lccccccc}
\toprule
\multirow{2}{*}{Method} & \multirow{2}{*}{\textbf{ZS}} & \multicolumn{2}{c}{\textbf{Val-Seen}} & \multicolumn{2}{c}{\textbf{Val-Seen-Syn}} & \multicolumn{2}{c}{\textbf{Val-Unseen}} \\
\cmidrule(lr){3-4} \cmidrule(lr){5-6} \cmidrule(lr){7-8}
&& SR$\uparrow$ & SPL$\uparrow$ & SR$\uparrow$ & SPL$\uparrow$ & SR$\uparrow$ & SPL$\uparrow$  \\
\midrule
BC    & \ding{55}      & 11.1 & 4.5  & 9.9  & 3.8  & 5.4  & 1.9 \\
DAgger~\cite{ross2011reduction} & \ding{55}        & 11.1 & 4.5  & 9.9  & 3.8  & 5.4  & 1.9 \\
RL~\cite{PPO}  & \ding{55}              & 18.1 & 9.4  & 15.0 & 7.4  & 10.2 & 4.7 \\
BCRL~\cite{wang2019reinforced}  & \ding{55}         & 39.2 & 18.7 & 27.8 & 11.7 & 18.6 & 7.5 \\
DAgRL~\cite{chen2019touchdown} & \ding{55}      & 41.3 & 21.2 & 29.4 & 14.4 & 18.3 & 7.9 \\
VLFM~\cite{yokoyama2024vlfm} &\checkmark & 35.2 & 18.6 & 32.4 & 17.3 & 35.2 & 19.6 \\
DAgRL+OD~\cite{yokoyama2024hm3d} & \ding{55} & 38.5 & 21.1 & 39.0 & 21.4 & 37.1 & 19.8 \\
Uni-NaVid~\cite{UniNavid:RSS25} & \ding{55} & 41.3 & 21.1 & 43.9 & 21.8 & 39.5 & 19.8 \\
MTU3D~\cite{zhu2025move}  & \ding{55}    & 55.0 & 23.6 & 45.0 & 14.7 & 40.8 & 12.1 \\ 
Nav-R1~\cite{liu2025nav} & \ding{55} & 58.4 & 26.3 & 48.1 & 23.1 & 42.2 & 20.1 \\
\midrule
\rowcolor{mylight} \textbf{LongNav-R1} & \checkmark &   \textbf{59.2} & \textbf{32.1} & \textbf{58.7} &\textbf{28.3} & \textbf{53.8} & \textbf{22.8}  \\
% \rowcolor{mylight} \textbf{LongNav-R1}$^*$ &   &  &  & &  &  \\
\bottomrule
\end{tabular}
}
\vspace{-4mm}
\label{tab:main_ovon}
\end{table}

\noindent\textbf{LongNav-R1 outperforms previous SOTAs on object-goal navigation benchmarks: HM3D V1, V2, and MP3D.} Tab.~\ref{tab:main_hm3d} compares LongNav-R1 against state-of-the-art object navigation methods using various observation modalities, including odometry, depth, and RGB. LongNav-R1 consistently outperforms existing baselines across all datasets, improving success rates to \textbf{76.0}\%, \textbf{83.7}\%, and \textbf{63.0}\% on HM3D V1, V2, and MP3D, respectively. We also observe that: i) LongNav-R1 (using RGB-only) surpasses methods that rely on full odometry and RGBD sensor setups. This demonstrates the effectiveness of our approach and suggests strong generalizability to real-world scenarios, where odometry and depth sensors are often noisy or range-limited. We attribute this robustness to the informative navigation cues provided by large model priors and the spatial understanding enriched by multi-turn causal reasoning; and ii) LongNav-R1 outperforms the prior end-to-end VLA model UniNavid~\cite{UniNavid:RSS25} on path efficiency by \textbf{7.2}\%. This superior efficiency stems from our RL training, which enables the agent to learn diverse behaviors to solve the navigation objective. In contrast, UniNavid uses SFT imitation learning, which is limited by the strict mimicry of human demonstrations and struggles to adapt efficiently to unseen testing environments.

\begin{table}[!t]
\footnotesize
\setlength{\tabcolsep}{3pt} 
    \centering
    \caption{\textbf{Compare to existing large-scale learning strategies for VLAs.} Results are reported on OVON Val-Unseen.}
    \vspace{-2mm}
    \label{tab:baseline}
    \begin{tabular}{cccccc}\toprule
    Method & Turn & Learning & GPUh & SR$\uparrow$ & SPL$\uparrow$\\\midrule
    UniNavid~\cite{UniNavid:RSS25} & Single-turn & Imitation learning & 1400 & 39.5   & 19.8\\% & 5.9M 
    Nav-R1~\cite{liu2025nav} & Single-turn & Critic-free RL & -  & 42.2  & 20.1 \\ % & 110K 
    StreamVLN~\cite{wei2025streamvln} & Multi-turn & Imitation learning & 1500 & 11.4  & 9.4 \\ % & 300K 
    SFT~\cite{RLHF} & Multi-turn & Imitation learning & 72 & 45.5   & 19.9 \\ % & 30k
    \midrule
    \textbf{LongNav-R1} & Multi-turn & Critic-free RL & 144 & 53.8   & 22.8\\ % & 34k
    % PPO & M & CB RL & 216 & 0.0 & 0.0 \\
    \bottomrule
    \vspace{-4mm}
\end{tabular}
\end{table} 

\begin{figure}[!t]
    \centering
    \includegraphics[width=0.48\linewidth]{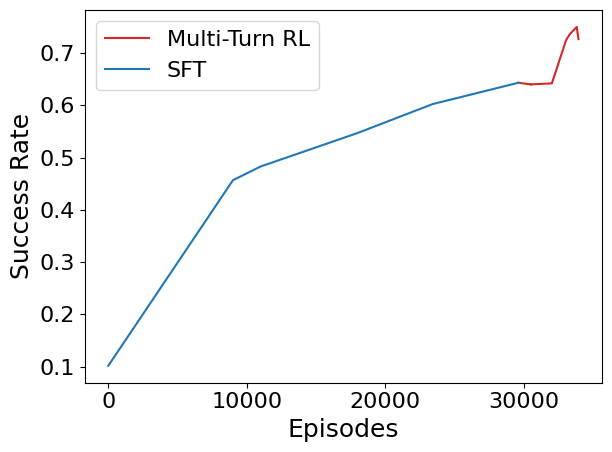}
    \includegraphics[width=0.48\linewidth]{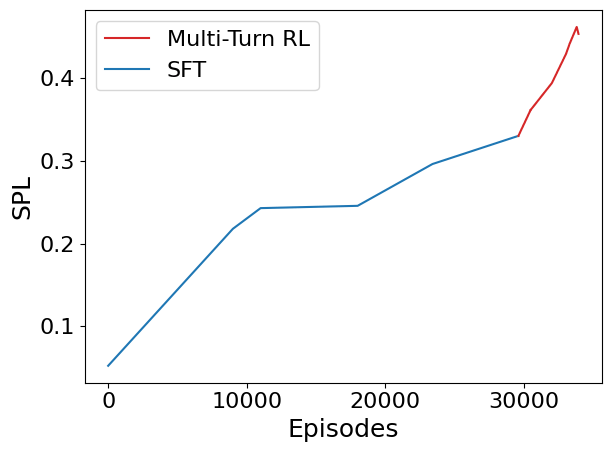}
    \vspace{-1mm}
    \caption{RL demonstrates high efficiency, improving performance by 30\% with 4k iterations. Evolution of success rate and path efficiency during training. The training pipeline consists of two phases: an initial SFT phase followed by a RL phase.}
    \label{fig:Infer}
    \vspace{-4mm}
\end{figure}
\noindent\textbf{LongNav-R1 outperforms previous SOTAs on the open-vocabulary object-goal navigation benchmark: HM3D-OVON.} Tab.~\ref{tab:main_ovon} compares LongNav-R1 against several state-of-the-art open-vocabulary object navigation methods. To validate the generalization ability of LongNav-R1, we trained exclusively on HM3D and evaluated on all other benchmarks via direct inference. LongNav-R1 demonstrates strong generalization: without any fine-tuning, it achieves SOTA. We also observe that: i) LongNav-R1 significantly surpasses traditional transformer-based RL methods by about 30\%, such as RL~\cite{PPO}, BCRL~\cite{wang2019reinforced}, and DAgRL~\cite{chen2019touchdown}. This highlights the potential of using pretrained large models as policies for robotic tasks; ii) LongNav-R1 outperforms the single-turn RL baseline, Nav-R1~\cite{liu2025nav}. We attribute this to our multi-turn RL approach, which better captures sequential dependencies and long-term navigation objectives.

\noindent\textbf{Evaluation against other large-scale learning strategies for VLAs.} We benchmarked against SOTA VLAs (UniNavid, Nav-R1) and RL baselines (REINFORCE++). LongNav-R1 excels across turn types and learning setups (critic-free RL and imitation learning). i) It surpasses multi-turn  (StreamVLN, SFT) by overcoming behavior cloning’s distribution shift and fostering better generalization, while StreamVLN fails to generalize. ii) Lacking existing critic-based RL for navigation VLAs, we reimplement PPO, however it collapses from value-head optimization complexities. LongNav-R1’s critic-free estimator removes this need, achieving superior performance and computational efficiency. 
% Note that we will extend HAPO to broader long-horizon tasks, e.g. mobile manipulation VLAs in future work. 

\noindent\textbf{Training statistics analysis.} Fig.~\ref{fig:Infer} shows the evolution of success rate and path efficiency throughout the training process, which consists of an initial SFT phase followed by RL. We see that: i) during the warm-up stage, SFT facilitates rapid early learning, elevating performance from 0 to 64.3\%; however, it encounters a bottleneck at approximately 30k iterations, where further data scaling yields diminishing returns; and ii) in contrast, the RL phase successfully breaks this plateau, demonstrating superior efficiency by improving performance by 8.7\% within only 4k iterations. This is attributed to the multi-turn design, which enables the model to learn from hundreds of decision-making points per sample, and the use of on-policy rollouts, which allow for effective error correction and behavioral refinement.

% \subsection{Evaluation on long-horizon navigation}

% \noindent\textbf{Comparison on outdoor long-horizon navigation.}

\begin{table}[!t]
\centering
\caption{\textbf{Effectiveness of multi-turn reinforcement learning.} Performance across different horizon length (steps). The physical distances are 12.5m (50 steps) and 50m (200 steps)}
% \footnotesize
\setlength{\tabcolsep}{2pt} 
\begin{tabular}{cccccccccc}
\toprule
\multirow{2}{*}{\textbf{Training Strategy}}& \multicolumn{2}{c}{\textbf{Overall}} & \multicolumn{2}{c}{\textbf{$<$ 50}} & \multicolumn{2}{c}{\textbf{50 - 200}}  & \multicolumn{2}{c}{\textbf{$>$ 200}} \\ 
\cmidrule(lr){2-3} \cmidrule(lr){4-5} \cmidrule(lr){6-7} \cmidrule(lr){8-9} 
 & SR$\uparrow$ & SPL$\uparrow$& SR$\uparrow$ & SPL$\uparrow$& SR$\uparrow$ & SPL$\uparrow$& SR$\uparrow$ & SPL$\uparrow$ \\ \midrule
% No Training & 0.51 & 0.50 & 0.54 & 0.52 & 0.00 & 0.00 & 0.00 & 0.00 & \\
% SFT~\cite{RLHF} & 64.32 & 32.99 & 72.97 & 50.71 & 70.08 & 34.13 & 34.29 & 10.13 & \\
No Training~\cite{yang2025qwen3} & 0.51 & 0.50 &1.9 & 1.8 & 0.0 & 0.0 & 0.0 & 0.0 \\
SFT~\cite{RLHF} & 64.3 & 33.0 &74.1 & 38.9 & 64.7 & 32.8 & 0.0 & 0.0& \\
REINFORCE++~\cite{hu2025reinforce++} & 47.4 & 26.0 & 68.0 & 37.0 & 42.5 & 23.5 & 0.0 & 0.0\\\midrule
\textbf{HAPO ($\sigma=\infty$)} & 71.6 & 40.4 & 82.7 & 47.3 & 72.2 & 40.5 & 0.0 & 0.0 \\
% Single-turn RL && && && & \\
\rowcolor{mylight}  \textbf{HAPO ($\sigma=30$)} & 73.0 & 44.3 & 86.2 & 53.2 & 69.9 & 41.7 & 15.4 & 9.4\\
\bottomrule
\end{tabular}
\label{tab:multiturn}
\vspace{-2mm}
\end{table}

% \begin{table}[th]
% \centering
% \caption{\textbf{Effectiveness of sparse global memory.} Performance across different horizon length.}
% % \footnotesize
% \setlength{\tabcolsep}{3pt} 
% \begin{tabular}{cccccccccc}
% \toprule
% \multirow{2}{*}{\textbf{Pruning}} & \multicolumn{2}{c}{\textbf{Overall}} & \multicolumn{2}{c}{\textbf{$<$ 50}} & \multicolumn{2}{c}{\textbf{50 - 200}}  & \multicolumn{2}{c}{\textbf{$>$ 200}} \\ 
% \cmidrule(lr){2-3} \cmidrule(lr){4-5} \cmidrule(lr){6-7} \cmidrule(lr){8-9} 
%  & SR$\uparrow$ & SPL$\uparrow$& SR$\uparrow$ & SPL$\uparrow$& SR$\uparrow$ & SPL$\uparrow$& SR$\uparrow$ & SPL$\uparrow$ \\ \midrule
% No sparse & \\
% Sparse 0.95 & \\
% Sparse 0.90 &  \\
% Sparse 0.85 &  \\
% \bottomrule
% \end{tabular}
% \label{tab:pruning}
% \vspace{-2mm}
% \end{table}

\begin{figure}[!t]
    \centering
    % --- First Subfigure ---
    \begin{subfigure}[b]{0.49\linewidth}
        \centering
        \includegraphics[width=\linewidth]{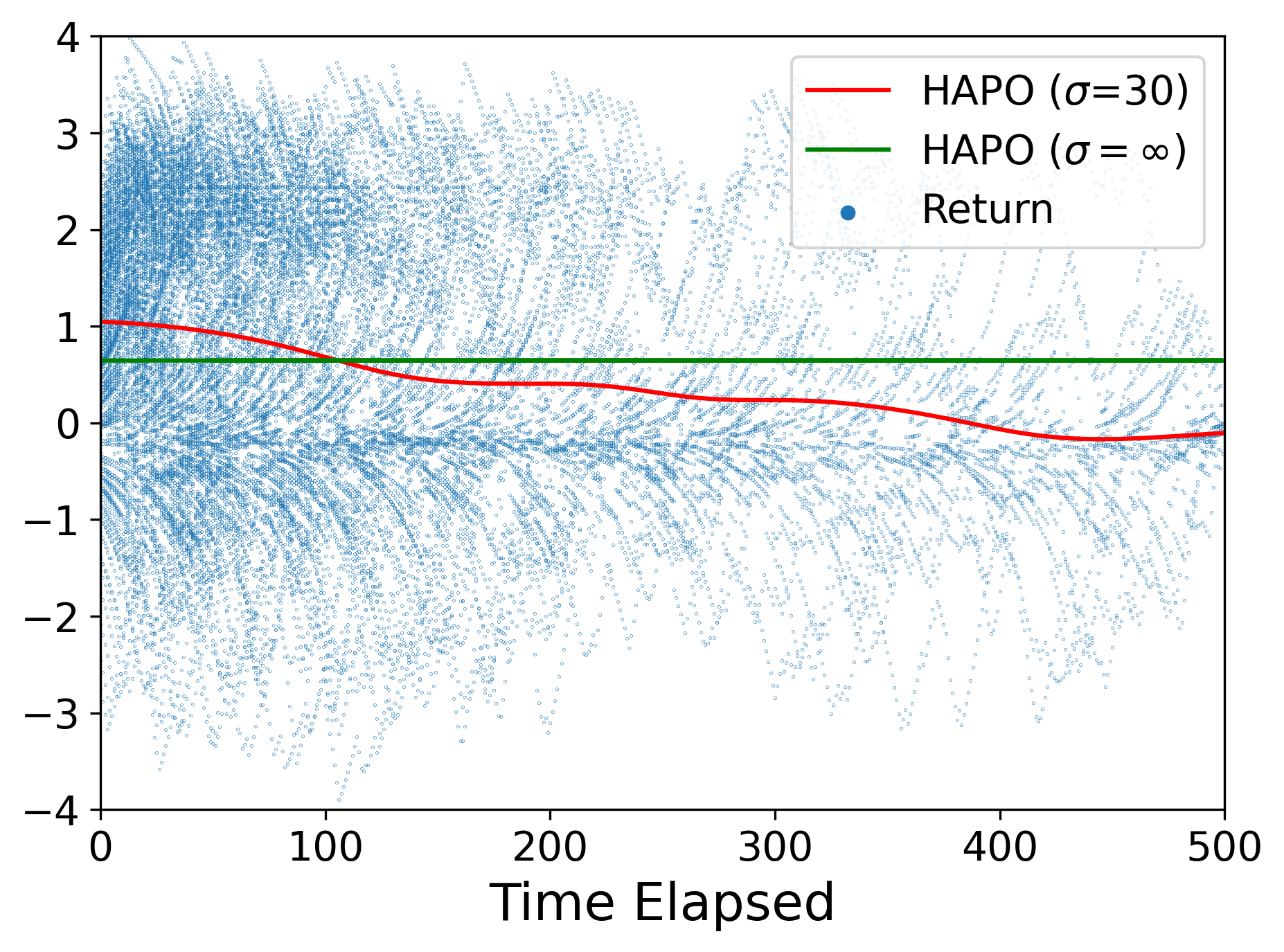}
        \vspace{-5mm}
        \caption{Baseline: Kernel $t$}
        \label{fig:baseline_te}
    \end{subfigure}
    \hfill 
    % --- Second Subfigure ---
    \begin{subfigure}[b]{0.49\linewidth}
        \centering
        \includegraphics[width=\linewidth]{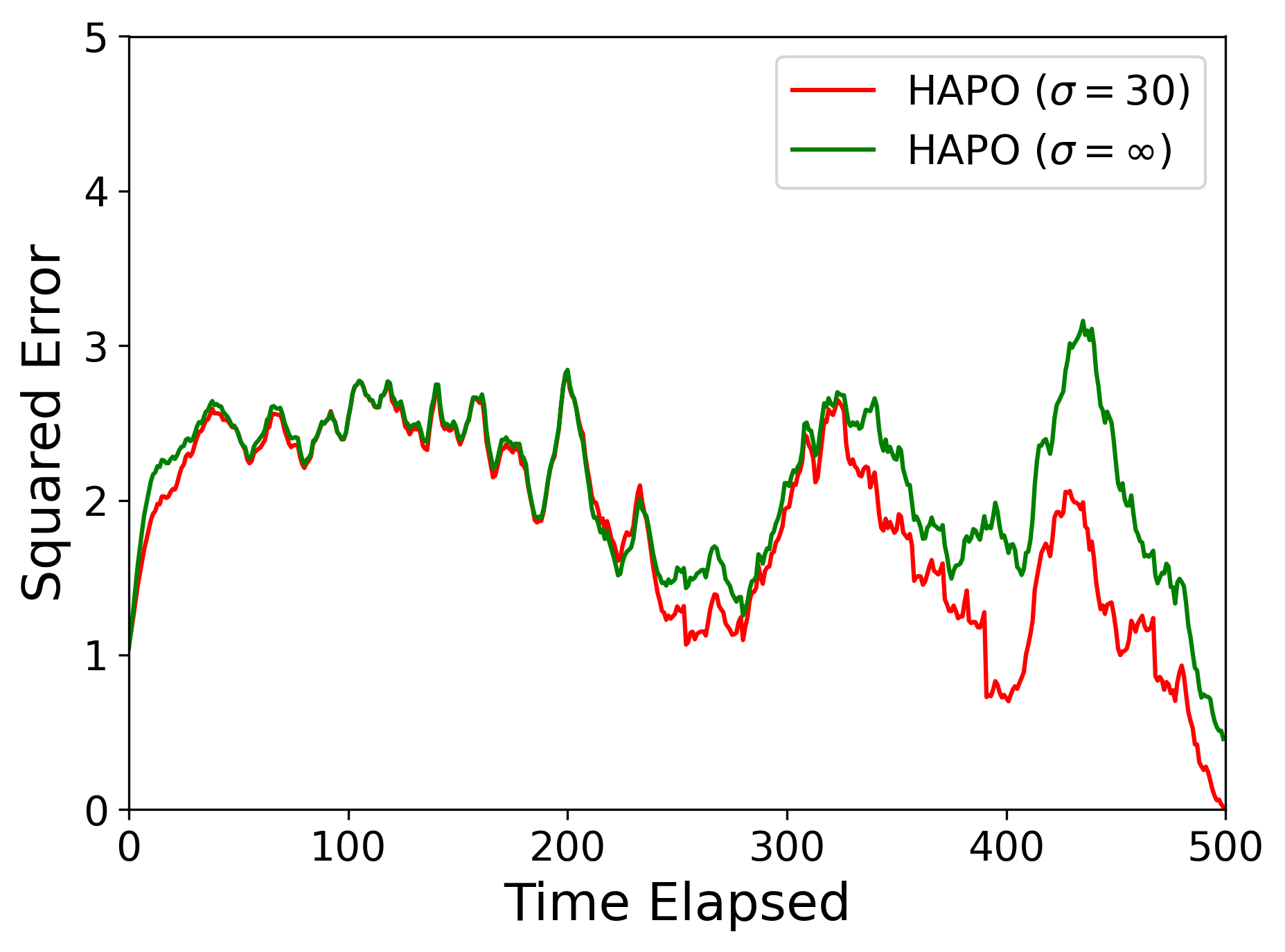}
        \vspace{-5mm}
        \caption{Val Err: Kernel $t$}
        \label{fig:error_te}
    \end{subfigure}
    % \hfill 
    % % --- Third Subfigure ---
    % \begin{subfigure}[b]{0.24\linewidth}
    %     \centering
    %     \includegraphics[width=\linewidth]{figs/time_to_go_fit.png}
    %     \vspace{-5mm}
    %     \caption{Baseline: Kernel $|\tau^i|-t$}
    %     \label{fig:baseline_tgo}
    % \end{subfigure}
    % \hfill 
    % \begin{subfigure}[b]{0.24\linewidth}
    %     \centering
    %     \includegraphics[width=\linewidth]{figs/time_to_go_err.png}
    %     \vspace{-5mm}
    %     \caption{Val Err: Kernel $|\tau^i|-t$}
    %     \label{fig:error_tgo}
    % \end{subfigure}
    \vspace{-5mm}
    \caption{We compare two HAPO variants, using different kernel sizes $\sigma$. Subplot (a) displays the returns in a batch buffer alongside the regressed baselines, while (b) illustrates the value estimation errors.}
    \label{fig:advantage_analysis}
    % \vspace{-1mm}
\end{figure}

\begin{figure}[!t]
    \centering
    % --- Left Side: The Figure ---
    \begin{minipage}{0.61\linewidth}
        \vspace{-3mm}
        \caption{Ablation of $\sigma$ and $\gamma$.}
        \vspace{-2mm}
        \resizebox{\linewidth}{!}{\includegraphics[width=1.0\linewidth]{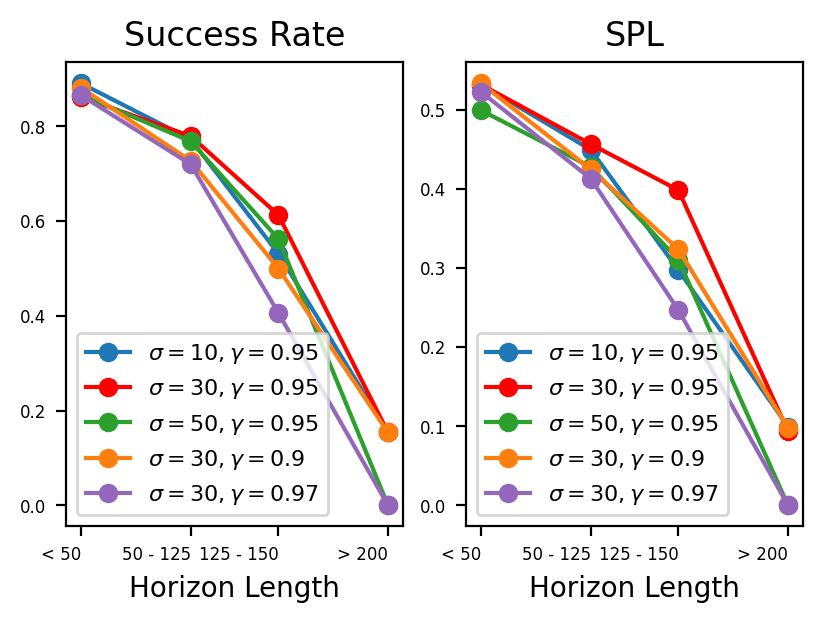}}
        % \captionof{table}{Ablation of bandwidth $\sigma$ and discount $\gamma$ \textcolor{blue}{Figure?}.}
        \vspace{-1mm}
        \label{tab:temporal}
    \end{minipage}\hfill % \hfill pushes them apart, the % prevents hidden spaces
    % --- Right Side: The Table ---
    \begin{minipage}{0.35\linewidth}
        \centering
        \vspace{-5mm}
        \footnotesize % Optional: shrinks table font if needed
        \setlength{\tabcolsep}{2pt}
        \captionof{table}{Feature.}
        \label{tab:feature}
        \vspace{-1mm}
        \resizebox{0.8\linewidth}{!}{\begin{tabular}{ccc}
            \toprule
            Feature & SR$\uparrow$ & SPL$\uparrow$ \\\midrule
            Timestep & 76.4 & 45.7  \\
            Per-Traj & 71.0  & 40.5 \\
            % Per-Traj (sparse)& 70.5 & 39.3  \\
            Distance & 73.4 & 45.3 \\\bottomrule
        \end{tabular}}
        \vspace{-1mm}
        \captionof{table}{Horizon.}
        \label{tab:horizon}
        \resizebox{\linewidth}{!}{\begin{tabular}{ccccc}
            \toprule
            % \multirow{2}{*}{$\sigma$} & \multicolumn{2}{c}{<200} & \multicolumn{2}{c}{>200}\\
            % \cmidrule(lr){2-3}  \cmidrule(lr){4-5}
            % & SR$\uparrow$ & SPL$\uparrow$ & SR$\uparrow$ & SPL$\uparrow$\\\midrule
            % $30$ & 75.0 & 45.3 & 15.4 & 9.4\\
            % $\infty$ & 75.1	& 42.4 & 0.0 & 0.0 \\
            \multirow{2}{*}{$\sigma$} & \multicolumn{2}{c}{<200} & \multicolumn{2}{c}{>200}\\
            \cmidrule(lr){2-3}  \cmidrule(lr){4-5}
            & VE $\downarrow$ & SR$\uparrow$ & VE $\downarrow$ & SR$\uparrow$ \\\midrule
            $30$ & 230 & 75.1 & 455 & 15.4 \\
            $\infty$ & 230 & 75.0 & 530 & 0.0 \\
            \bottomrule
        \end{tabular}}
    \end{minipage}
    \vspace{-7mm}
\end{figure}

\vspace{-2mm}
\subsection{Ablation studies}

\noindent\textbf{Effectiveness of multi-turn reinforcement learning.} Tab.~\ref{tab:multiturn} compares two HAPO variants against various training strategies, including zero-shot (no training), SFT~\cite{RLHF}, and REINFORCE++~\cite{hu2025reinforce++}.  Specifically, REINFORCE++ uses sparse binary outcome rewards (1/0 for success/failure), while HAPO incorporates step-wise distance-to-goal rewards. We see that: i) both HAPO variants outperform all other training strategies; and ii) RL with dense rewards outperforms SFT, whereas RL with sparse rewards does not. This demonstrates the importance of dense reward shaping in multi-horizon robotics tasks.

\noindent\textbf{Effectiveness of horizon-adaptive advantage estimation.} Fig.~\ref{fig:advantage_analysis} and Tab.~\ref{tab:multiturn} compares HAPO variants, using different bandwidth sizes $\sigma$. We visualize the estimated baseline and the value estimation error of a randomly sampled batch buffer in subplots (a) and (b). The value estimation error is the average difference between the real return and the regressed baseline at each step. HAPO with restricted temporal window outperforms infinite-window (uniform) baseline. We see that: i) HAPO with $\sigma=30$ can achieve smaller estimation error compared to $\sigma=\infty$; ii) this reduced estimation error directly correlates with improved performance, as evidenced in Tab.~\ref{tab:multiturn}. This highlights the necessity of temporal-adaptive advantage estimation, as simple uniform dense estimation fails to account for the complex temporal dynamics inherent in robotic navigation.

\noindent\textbf{Ablation on kernel bandwidth $\sigma$ and discount factor $\gamma$, see Fig.~\ref{tab:temporal}.} We see that an intermediate temporal window ($\sigma = 30, \gamma = 0.95$) yields optimal performance. This balance is critical because increasing temporal window via higher $\gamma$ prevents short-sighted policies by encouraging long-term predictions, yet setting it too high makes return values unpredictable. Increasing temporal window via higher $\sigma$ prevents the baseline from overfitting to local returns, but an excessively large $\sigma$ eliminates horizon awareness. We optimize this configuration via value-error analysis on an offline rollout dataset (Fig. 4 in main paper). This kernel regression is computationally efficient and strongly correlates with final model performance. Moreover, this technique can be extended to automated bandwidth tuning using an online rollout buffer.

\noindent\textbf{Stronger justification for timestep feature design, see Tab.~\ref{tab:feature}}. HAPO is a flexible framework capable of accommodating different factors via feature switching. Specifically, HAPO's timestep feature outperforms per-trajectory and distance metrics. The reasons are: i) HAPO reduces per-trajectory bias (similar to REINFORCE++ v.s GRPO), ensuring generalized navigation over instance-specific overfitting; ii) distance fails to credit critical in-place actions (turning to unlock regions). This is reflected in the data: elapsed timesteps maintain a higher correlation with returns ($0.24$) than distance-to-goal ($0.09$).

\noindent\textbf{Clearer explanation for value-error analysis, see Tab.~\ref{tab:horizon}.} i) Navigation success requires minimizing accumulated value error across all elapsed timesteps. ii) Thus, longer horizons (>200 steps) yield larger accumulated errors and lower performance than shorter ones (<200 steps). iii) For >200 steps, HAPO (30) sustains lower error and outperforms the $\infty$-baseline, while both perform identically for <200 steps. iv) The peak variations (<200) in Tab. III (main paper) arise from fine-grained binning across unbalanced intervals (<50,  50-200).

% and ii) utilizing time-to-go as a kernel feature achieves a lower estimation error (1.43) compared to time-elapsed (2.13). Since returns are inherently determined by future trajectories, grouping them by time-to-go aligns more effectively with the actual realized return. While time-elapsed represents an absolute temporal state, time-to-go serves as a normalized proxy for task difficulty, allowing the baseline to group states with similar reward potential. This approach effectively trades a higher bias for a significant reduction in variance.

\noindent\textbf{Computation analysis.} Tab.~\ref{tab:computation} compares the computational overhead of LongNav-R1 in terms of inference latency and memory consumption. We see that while inference time scales with the horizon, LongNav-R1 remains highly performant, requiring only 0.23 s to process a 400-step sequence. Achieving about 5 FPS, the model can be deployed in practical real-world environments. This is attributed to two architectural advantages: the multi-turn setup, which enables KV cache reuse to avoid redundant processing of past observations, and online token pruning, which restricts attention mechanisms to only the most informative new tokens.

\begin{table}[!t]
\centering
\caption{Computation analysis.}
% \footnotesize
\setlength{\tabcolsep}{3pt} 
\begin{tabular}{ccccccccc}
\toprule
\multirow{2}{*}{Turns} & \multicolumn{4}{c}{\textbf{Inference Time (s)}} & \multicolumn{4}{c}{\textbf{Memory Cost (GB)}} \\
\cmidrule(lr){2-5} \cmidrule(lr){6-9} 
& 50 & 100 & 200  & 400 & 50 & 100 & 200  & 400 \\\midrule
% Single-Turn &  \\
\rowcolor{mylight}  \textbf{LongNav-R1} & 0.12 & 0.15 & 0.19 & 0.23 & 6.25 & 7.34 & 9.33 & 11.62 \\
\bottomrule
\end{tabular}
\label{tab:computation}
\vspace{-3mm}
\end{table}

\begin{table}[!t]
\centering
\caption{Ablation of pruning.}
% \footnotesize
% \setlength{\tabcolsep}{3pt} 
\begin{tabular}{ccccc}
        \toprule
        Pruning ($\delta$) & Speed (s) & Mem (GB) & SR$\uparrow$ \\ \midrule
        0\% (1.0) & 0.203 & 15.6 & 53.1 \\
        20\% (0.97) & 0.172 & 13.7 & 56.5 \\
        35\% (0.95) & 0.168 & 11.8 & 54.4 \\
        45\% (0.93) & 0.141 & 10.8 & 52.4 \\
        50\% (0.92) & 0.132 & 10.4 & 45.6 \\
        63\% (0.90) & 0.118 & 8.9 & 25.8 \\
        80\% (0.85) & 0.089 & 7.3 & 0.0 \\
        \bottomrule
        \end{tabular}
        \label{tab:pruning}
\vspace{-3mm}
\end{table}

\noindent\textbf{Effectiveness of online pruning.} Tab.~\ref{tab:pruning} shows that increasing the pruning ratio consistently improves speed and reduces memory usage, navigation accuracy follows an inverted-U trend. Performance suffers at both extremes: unpruned models quickly exceed memory limits in long tasks (peak memory), while over-pruning discards essential data. ii) We demonstrate superior efficiency compared to baselines (StreamVLN requires 24GB memory and 0.37s). Moreover, performance drops after 150 steps (37.5m) due to the training focus on nearby targets and context-heavy degradation at 60k tokens, see Fig.\ref{tab:temporal}.

% \noindent\textbf{Effectiveness of sparse global memory.}

% \input{table/r2r}

% \input{table/outdoor}

\vspace{-2mm}
\subsection{Case study}

\begin{figure}[!t]
    \centering
    \begin{subfigure}[b]{0.85\linewidth}
        \centering
        \includegraphics[width=0.48\linewidth]{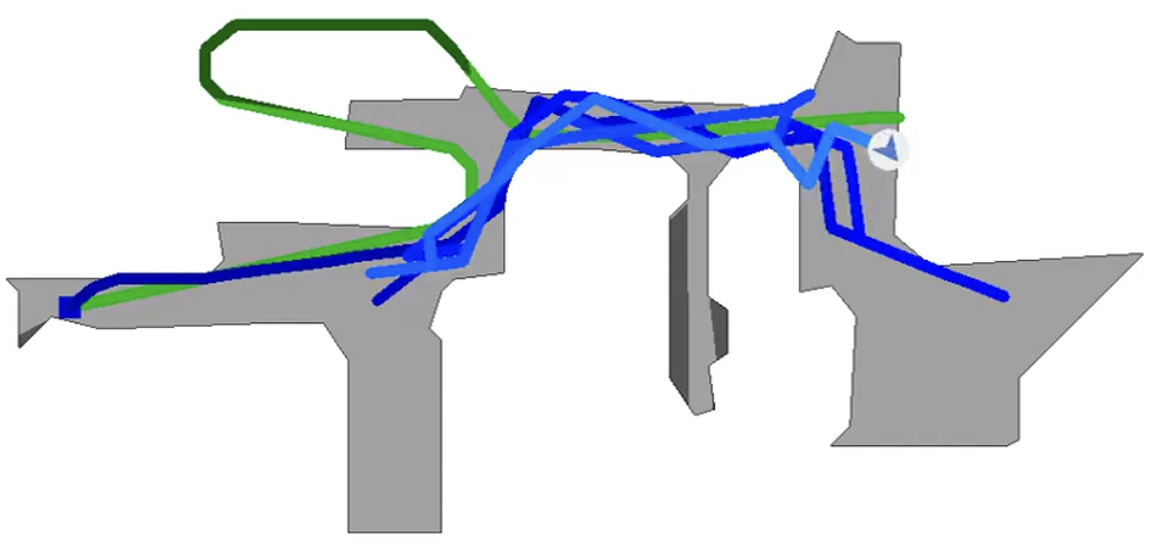}
        \includegraphics[width=0.35\linewidth]{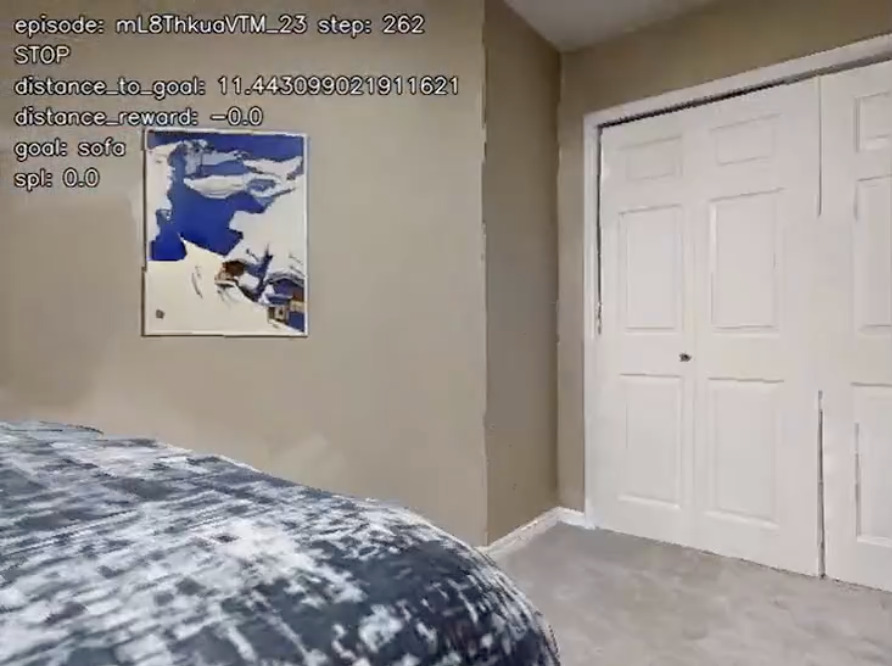}
        \vspace{-1mm}
        \caption{SFT}
        \label{fig:case_sft}
    \end{subfigure}
    \hfill 
    \begin{subfigure}[b]{0.85\linewidth}
        \centering
        \includegraphics[width=0.48\linewidth]{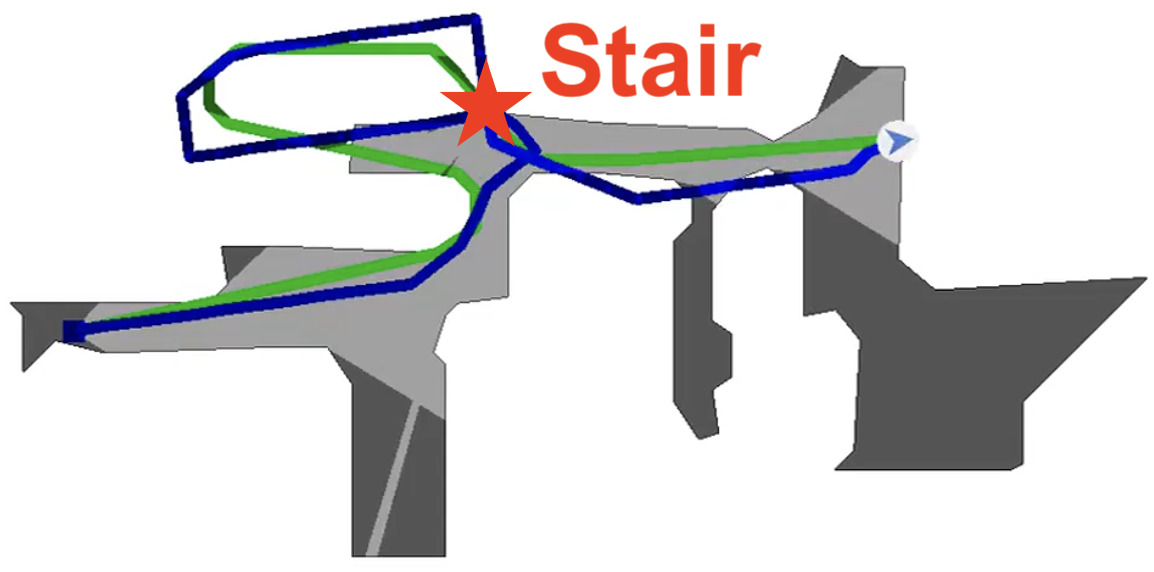}
        \includegraphics[width=0.35\linewidth]{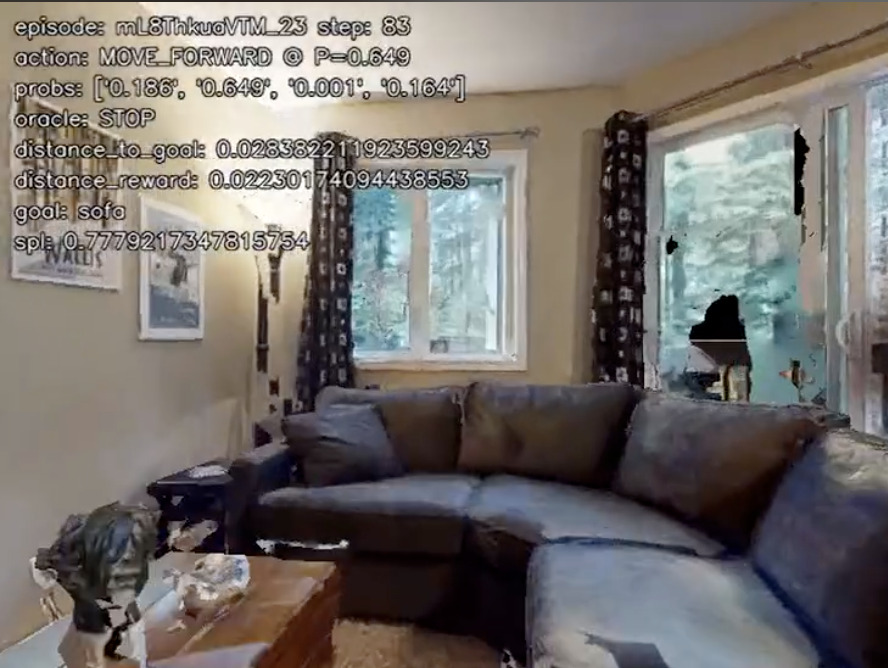}
        \vspace{-1mm}
        \caption{LongNav-R1}
        \label{fig:case_hapo}
    \end{subfigure}
    \vspace{-1mm}
    \caption{Visualization of the navigation process in habitat. \textcolor{blue}{Blue} and \textcolor{green}{green} lines represent the optimal and inferred trajectories.}
    \label{fig:case_study_habitat}
    \vspace{-3mm}
\end{figure}

\noindent\textbf{Qualitative analysis of SFT and LongNav-R1.} Fig.~\ref{fig:case_study_habitat} visualizes the navigation process of SFT and LongNav-R1 as they search for a sofa. We see that: i) despite the challenge of a distant, cross-floor goal, LongNav-R1 successfully and efficiently reaches the target while SFT fails; and ii) LongNav-R1 natively handles cross-floor transitions, a task where modular designs often struggle, enhancing its adaptability in real-world environments.

\begin{figure}[!t]
    \centering
    \includegraphics[width=\linewidth]{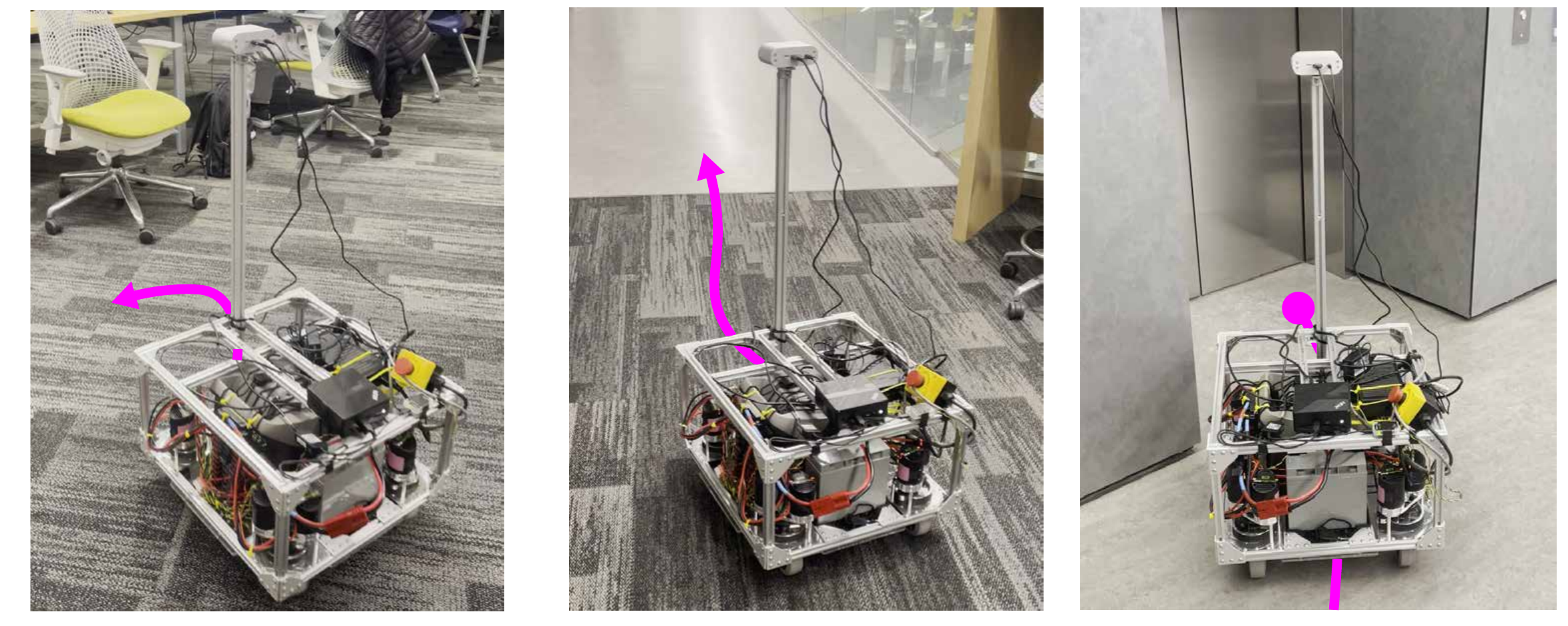}
    \vspace{-4mm}
    \caption{Visualization of the navigation process in \textbf{real-world}. The robot successfully navigates to an elevator, an unseen category. Displayed trajectories are approximate for visualization.}
    \label{fig:case_study_real}
    \vspace{-7mm}
\end{figure}

\noindent\textbf{Visualization of LongNav-R1 in real-world deployment.} Fig.~\ref{fig:case_study_real} visualizes the navigation process of LongNav-R1 as it searches for an elevator, an unseen object category, in a real-world office environment. For the hardware mobile base, we adopt the mobile base design from TidyBot++~\cite{wu2024tidybot}, omitting the manipulator. Equipped with LongNav-R1, the robot successfully locates the target, demonstrating zero-shot real-world effectiveness of the navigation policy.

% \begin{figure*}[!t]
%     \centering
%     \begin{subfigure}[b]{0.48\linewidth}
%         \centering
%         \includegraphics[width=0.24\linewidth]{figs/habitat_vis/SFT_BEV.jpg}
%         \includegraphics[width=0.48\linewidth]{figs/habitat_vis/SFT_End.jpg}
%         \includegraphics[width=0.48\linewidth]{figs/habitat_vis/RL_BEV.jpg}
%         \includegraphics[width=0.48\linewidth]{figs/habitat_vis/RL_End.jpg}
%         \vspace{-2mm}
%         \caption{SFT}
%         \label{fig:case_sft}
%     \end{subfigure}
%     \hfill 
%     \begin{subfigure}[b]{0.48\linewidth}
%         \centering
%         \includegraphics[width=0.48\linewidth]{figs/time_to_go_err.png}
%         \includegraphics[width=0.48\linewidth]{figs/time_to_go_fit.png}
%         \vspace{-2mm}
%         \caption{LongNav-R1}
%         \label{fig:case_hapo}
%     \end{subfigure}
%     \vspace{-1mm}
%     \caption{Visualization of the navigation process.}
%     \label{fig:case_study}
%     \vspace{-5mm}
% \end{figure*}
% LongNav-R1 successfully reaches a distant, cross-floor goal, whereas SFT terminates prematurely. This highlights SFT's inability to generalize to long-horizon tasks due to imitation bias from shorter human demonstrations.
\vspace{-1mm}
\section{Discussion}
\vspace{-1mm}
\noindent\textbf{Generalize to tasks lacking geometric proxies.} i) We want to clarify that HAPO is not constrained to dense geometric proxies. As a versatile, critic-free framework for long-horizon tasks, it accommodates diverse factors via feature switching and natively supports sparse reward (producing the REINFORCE++ in Tab.~\ref{tab:multiturn}); ii) HAPO’s strong sim-to-real generalization, evidenced by its superior zero-shot performance in real-world demo, makes it highly feasible to get these priors in simulation before deploying in real world.

% \noindent\textbf{Adapt to dynamic settings.} LongNav-R1 can automatically adapt to dynamic settings via real-time memory updates to incorporate fresh environmental observations. 

\vspace{-1mm}
\section{Conclusion}
\vspace{-1mm}

\noindent\textbf{Conclusion.} In this paper, we provide a recipe for training end-to-end VLA policies for long-horizon navigation using multi-turn reinforcement learning and horizon-adaptive policy optimization. By utilizing a critic-free advantage estimation, HAPO effectively manages dense rewards and sequential decision-making without the prohibitive computational overhead of an auxiliary critic model, a key advantage when scaling large models. Our model, LongNav-R1, achieves state-of-the-art results across four simulation benchmarks and demonstrates robustness in real-world deployment. 

% Furthermore, we demonstrate that LongNav-R1 maintains high learning and computational efficiency even over extremely long horizons, proving its viability for deployment in complex, real-world environments.

\noindent\textbf{Limitation and future works.} A current limitation is that our online token pruning retains the full KV cache to maintain causal consistency. To mitigate memory bottlenecks in extreme-horizon tasks, future work will explore selective KV cache eviction and spatially aware global memory mechanisms. Beyond navigation, we aim to extend this framework to diverse applications, such as mobile manipulation, and incorporate world models for real-world multi-turn reinforcement learning.

\clearpage
{
    \small
    \bibliographystyle{plainnat}
    \bibliography{references}
}
\clearpage
\section{Appendix}

% \subsection{Formulation}

\subsection{SFT implementation details}

\begin{figure*}[!ht]
    \centering
    \includegraphics[width=0.92\linewidth]{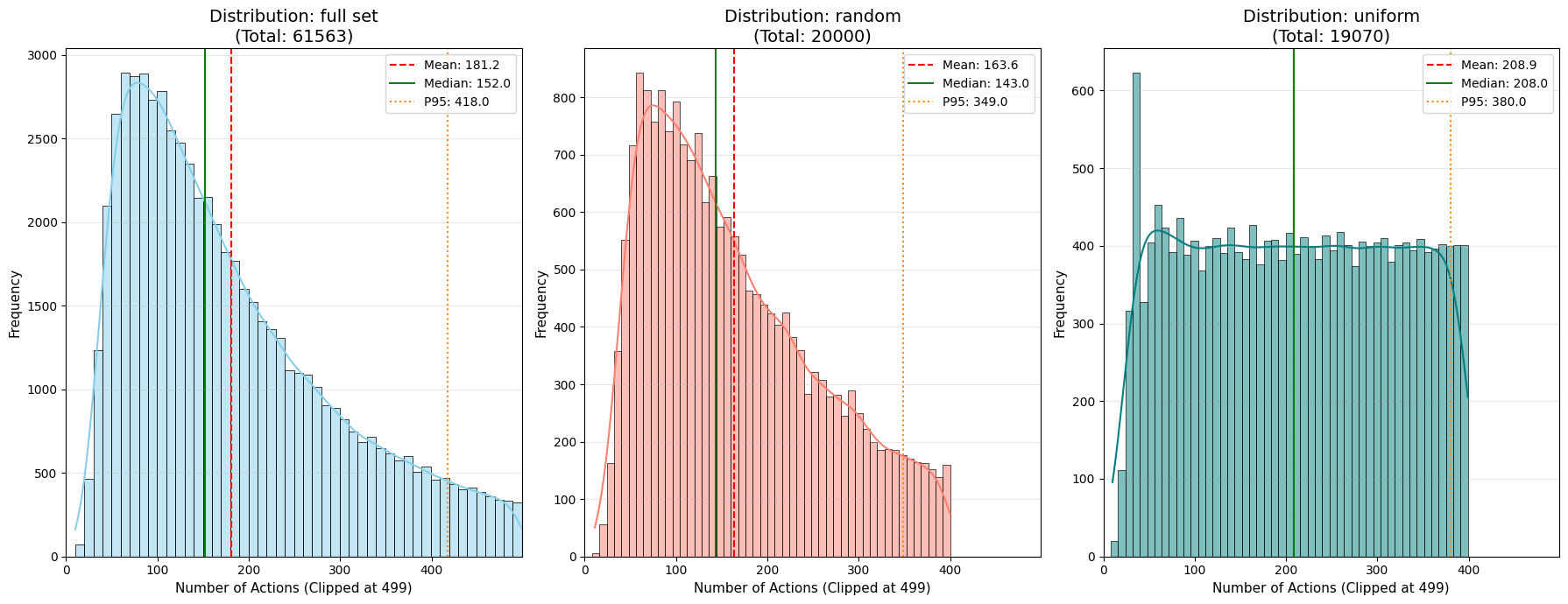}
    \vspace{-2mm}
    \caption{Visualization of SFT trajectory data distribution. (a) is the overall human demonstration data, (b) is the randomly sampled data distribution, and (c) is our uniformly sampled data distribution.}
    \label{fig:sft_data}
    \vspace{-5mm}
\end{figure*}

\noindent\textbf{SFT Data Curation.} To bootstrap the VLM for the navigation objective, we perform Supervised Fine-Tuning (SFT) utilizing human demonstrations from the Habitat-Web dataset~\cite{Habitatweb}. To ensure training efficiency and accommodate GPU memory limitations, we truncated the raw human trajectory data at 500 steps. We then evaluated various sampling strategies to effectively utilize the data. From Fig.~\ref{fig:sft_data}, we can see that the dominant trajectory length is approximately 100 steps. Experimentally, we see that naive random sampling from this distribution introduces a significant ``stop bias,'' causing the model to over-fit to premature episode termination. Consequently, scaling randomly sampled data leads to performance degradation; as shown in Tab.~\ref{tab:sft_data}, the Success rate drops from 55.5\% at 4k samples to 46.5\% at 30k samples, despite improvements in Success weighted by Path Length.

To mitigate this, we implement a Uniform Sampling strategy that equalizes trajectory selection across lengths ranging from 0 to 400 steps. This approach balances the state-action space, ensuring robust policy learning across varying trajectory complexities. As demonstrated in Tab.~\ref{tab:sft_data}, uniform curation prevents convergence on sub-optimal stop behaviors and effectively leverages larger datasets. At the 30k data scale, the uniform configuration achieves a peak Success rate of 64.3\% and a Success SPL of 57.0, representing a 17.8\% absolute increase in success over the random baseline.

\begin{table}[!ht]
\centering
\caption{Effectiveness of SFT data curation.}
\begin{tabular}{llccc}
\toprule
\textbf{Sampling} & \textbf{Data} & \textbf{Success} & \textbf{SPL} & \textbf{Success SPL} \\
\midrule
Random & 4k & 55.5 & 25.0 & 45.0 \\
Random & 10k & 49.5 & 25.6 & 51.8 \\
Random & 30k & 46.5 & 24.7 & 53.1 \\
Uniform & 10k & 53.0 & 27.6 & 52.0 \\
Uniform & 30k & 64.3 & 33.0 & 57.0 \\
\bottomrule
\end{tabular}
\label{tab:sft_data}
\end{table}

\noindent\textbf{SFT training.}
We use Qwen3-VL 2B Instruct~\cite{yang2025qwen3} as the base model and employ LoRA~\cite{hu2021loralowrankadaptationlarge} for parameter efficient SFT fine tuning. LoRA adapters with configuration $\{r=128, \alpha=256, 
\text{dropout}=0.05\}$ are attached to all linear blocks of the language model, while the vision encoder is left frozen.

We train with an initial learning rate of $3 \times 10^{-5}$ and a linear decaying schedule. We configure the AdamW optimizer with the standard hyperparameters $\{\beta_1=0.9,\beta_2=0.999\}$. Gradient checkpointing along with our visual token sparsification enables bfloat16 training on full trajectories up to 400 steps long, without need for quantization or model parallelism. Training completes in 18 hours on 2 H100 GPUs.
\subsection{RL implementation details}
\noindent\textbf{Environment Setup.} Our RL training uses the 80 scene training split from the 2022 Habitat ObjectNav challenge~\cite{habitatchallenge2022}. We use the dense shaped reward $r_t(a_t,s_t)=-\Delta_{\rm geo\_dist}-0.01$ and terminal reward $r_T=2.5\cdot \text{SPL}$ standard to habitat ObjectNav tasks ~\cite{wijmans2019dd}. The RGB observations and cylindrical agent form factor are likewise habitat ObjectNav defaults. We set the maximal environment steps per episode to 350 to reduce computational cost. 

% As is typical for end-to-end methods ~\cite{zhang2024uninavid}\cite{cai2024pixnav}, we enable agent sliding against obstacles. 

% \begin{figure}[!t]
%     \centering
%     % \includegraphics[width=0.24\linewidth]{figs/habitat_vis/SFT_BEV.jpg}
%     % \includegraphics[width=0.24\linewidth]{figs/habitat_vis/SFT_End.jpg}
%     % \includegraphics[width=0.24\linewidth]{figs/habitat_vis/SFT_End.jpg}
%     % \includegraphics[width=0.24\linewidth]{figs/habitat_vis/SFT_End.jpg}
%     % \includegraphics[width=\linewidth]{figs/longnav.pdf}
%     \fbox{\parbox{0.92\linewidth}{\vspace{0.7in}\centering Comparison of step distribution of SFT and RL.}}
%     \vspace{-2mm}
%     \caption{Visualization of RL and SFT step distribution.}
%     \label{fig:sft_data}
%     \vspace{-5mm}
% \end{figure}

\noindent\textbf{Multi-turn RL framework.}
We implement a synchronous RL framework similar to DD-PPO~\cite{wijmans2019dd}, where experience collection is followed by training with gradient synchronization via torch DDP. However, instead of using a preemption threshold as in DD-PPO, we enable each worker to collect multiple trajectories per cycle using an asynchronous dispatch loop that pairs each VLM instance with dedicated Habitat Sim workers, amortizing the computation cost from episode length variability. Specifically, we dispatch episodes to any idle workers until the number of dispatches equals the rollout buffer size. This improvement allows us to use every single collected experience instead of discarding incomplete trajectories. Furthermore, VLM idle time is minimized as each VLM continuously runs episodes until the number of completed and running episodes equals the rollout buffer size.
%this is not true since it's asynchronous can we call real quick
% do not define new symbol without explaining it. ok move on.

\noindent\textbf{Multi-turn RL training.} To stabilize the RL training with limited computation resource, we use a rollout buffer size of 16 but retain the most recent 256 trajectories for HAPO advantage estimation. This design decision trades trajectory staleness when estimating baselines for reduced variance. 
% We found this tradeoff to be acceptable as training updates are only done on the 16 on policy trajectories.

We employ the dual clip PPO training objective in combination with ClipCov~\cite{cui2025entropy} to prevent entropy collapse. The PPO objective is configured with $\{\gamma =0.95,\rm clip\_ratio\_high=0.28, \rm clip\_ratio\_low=0.2\}$. ClipCov is configured with the standard hyperparameters $\{\rm  clip\_cov\_ub=5,\rm  clip\_cov\_lb=1,\rm clip\_cov\_ratio=0.0002\}$. We further regularize the policy with a KL divergence constraint. Since we train with LoRA adapters, disabling the adapters gives us the SFT tuned base model, which we use as the reference model. Since the habitat navigation action space is much smaller than typical large language model vocabulary size, we are able to calculate the full KL divergence instead of using KL estimators typically used in LLM/LVLM reinforcement learning. We use $\rm kl\_coeff=0.001$ to weigh the KL divergence from the SFT model.

Our RL policy was trained for 18 hours on 4 A100 GPUs, experiencing approximately 4000 episodes. Due to the need to accommodate long trajectories, each GPU runs a single VLM instance with batch size 1 during both rollout collection and training. Since the vast majority of trajectories are short enough to occupy less than half the GPU VRAM, we anticipate future performance gains by implementing dynamic batching.

\subsection{Online token pruning algorithm}
Our online visual token pruning strategy is consistent across both SFT and RL phases and is fully causal, allowing for efficient single-pass trajectory training. This approach enables the entire trajectory to be processed in one forward and backward pass, significantly boosting training efficiency.

\begin{figure*}[!t]
\centering
    \includegraphics[width=0.24\linewidth]{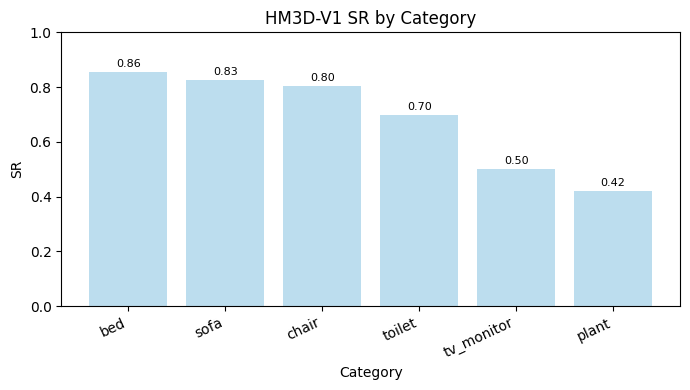}
    \includegraphics[width=0.24\linewidth]{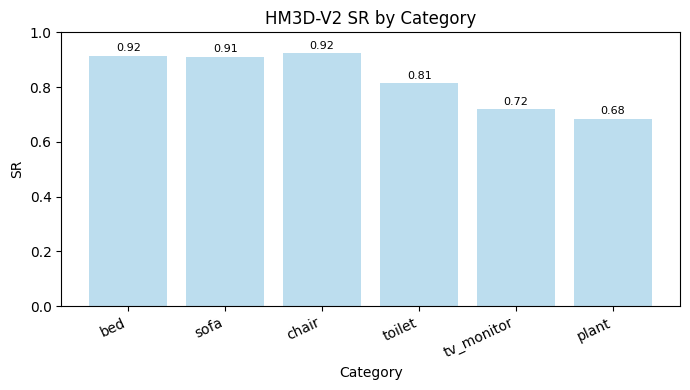}
    \includegraphics[width=0.24\linewidth]{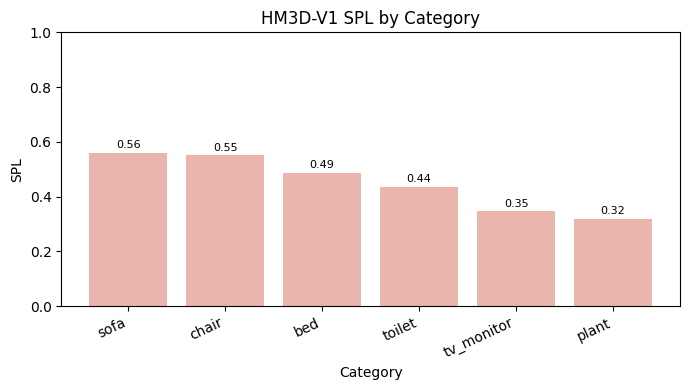}
    \includegraphics[width=0.24\linewidth]{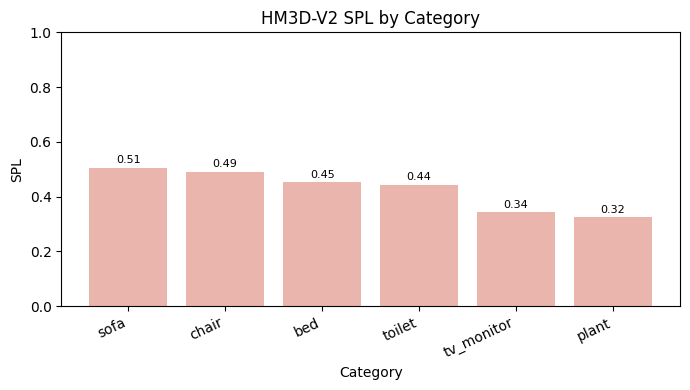}
    \includegraphics[width=0.48\linewidth]{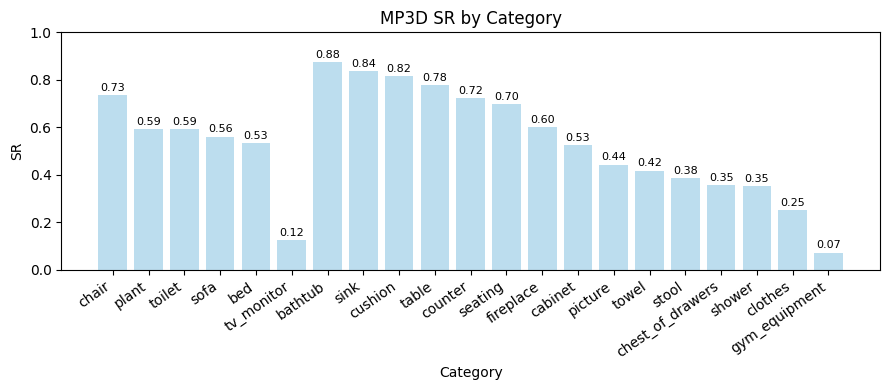}
    \includegraphics[width=0.48\linewidth]{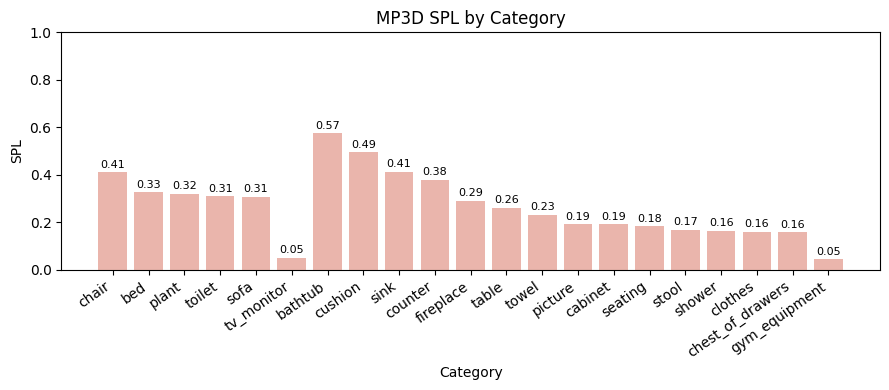}
    \includegraphics[width=0.75\linewidth]{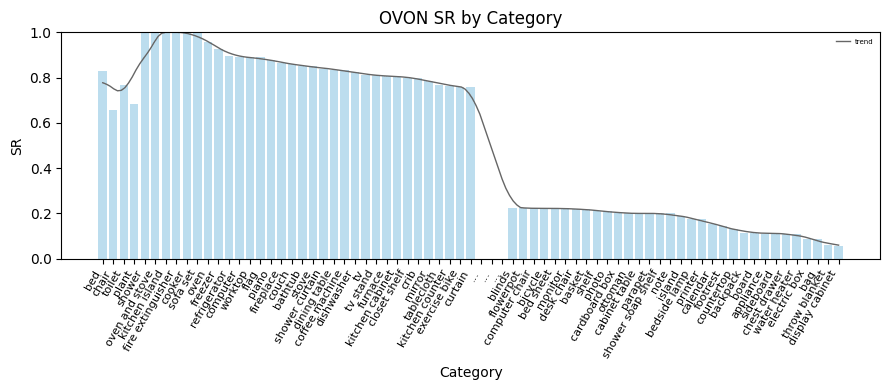}
    \includegraphics[width=0.75\linewidth]{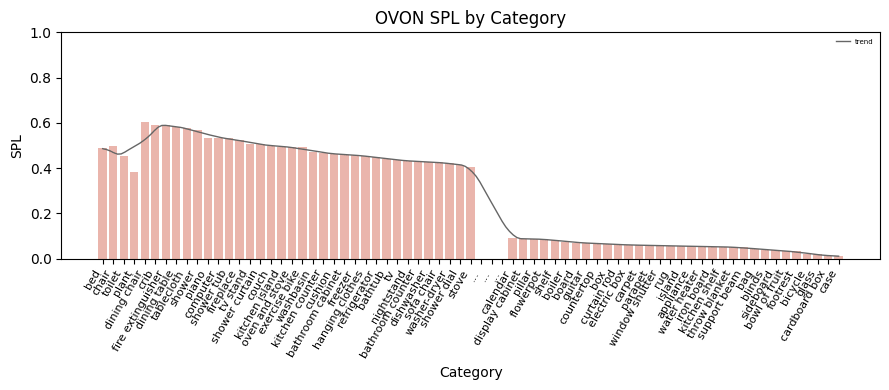}
    \vspace{-1mm}
    \caption{Performance across the 6 object categories in HM3D V1, V2, MP3D and OVON.}
    \label{fig:category}
    \vspace{-3mm}
\end{figure*}

% \begin{figure}[!t]
% \centering
%     \includegraphics[width=0.48\linewidth]{figs/hm3d-v1/hm3d-v1-sr.png}
%     \includegraphics[width=0.48\linewidth]{figs/hm3d-v2/hm3d-v2-sr.png}
%     \includegraphics[width=1.0\linewidth]{figs/mp3d/mp3d-sr.png}
%     \includegraphics[width=0.48\linewidth]{figs/hm3d-v1/hm3d-v1-spl.png}
%     \includegraphics[width=0.48\linewidth]{figs/hm3d-v2/hm3d-v2-spl.png}
%     \includegraphics[width=1.0\linewidth]{figs/mp3d/mp3d-spl.png}
%     \vspace{-1mm}
%     \caption{Performance across the 6 object categories in HM3D V1, V2 and MP3D.}
%     \label{fig:category}
%     \vspace{-3mm}
% \end{figure}

% \begin{figure*}[!t]
% \centering
%     \includegraphics[width=0.95\linewidth]{figs/ovon/ovon-sr.png}
%     \includegraphics[width=0.95\linewidth]{figs/ovon/ovon-spl.png}
%     \vspace{-1mm}
%     \caption{Performance across the 6 object categories in OVON.}
%     \label{fig:category_ovon}
%     \vspace{-3mm}
% \end{figure*}

\begin{table}[!ht]
  \centering
  \caption{Object categories of HM3D V1~\cite{hm3dv1}, V2~\cite{hm3dv2}, MP3D~\cite{Matterport3D} and HM3D-OVON~\cite{yokoyama2024hm3d}.}
  \small
  \setlength{\tabcolsep}{3pt}
  \begin{tabular}{lll}
    \toprule
    \textbf{Dataset} & \textbf{Object} & \textbf{Category} \\
    \midrule
    HM3D & 6 & Bed, sofa, tv\_monitor, plant, chair, toilet \\
    \midrule
    \multirow{4}{*}{MP3D} & \multirow{4}{*}{21} & Bed, sofa, tv\_monitor, plant, chair, toilet, \\
    && cabinet, chest\_of\_drawers, table, picture, sink, \\
    && stool, towel, cushion, shower, bathtub, counter,\\
    && fireplace, gym\_equipment, seating, clothes \\
    \midrule
    \multirow{20}{*}{OVON} & \multirow{20}{*}{379} & Bed, plant, chair, toilet, air\_conditioner, amplifier, \\
    && antique\_clock, antique\_telephone, appliance, \\
    && arcade\_game, archway, armchair, artwork, \\
    && baby\_changing\_station, backpack, backrest, \\
    && backsplash, bag, balcony\_railing, balustrade, \\
    && banner, bar, bar\_cabinet, bar\_chair, barbecue, \\
    && basket, bath\_cabinet, bath\_sink, bath\_towel, \\
    && bathrobe, bathroom\_accessory, bathroom\_cabinet,\\
    && \dots,\\
    && stage, stair, staircase\_handrail, staircase\_trim, \\
    && stairs\_railing, stand, statue, stereo\_set, \\
    && stone\_support\_structure, stool, storage, \\
    && storage\_cabinet, storage\_space, stove, \\
    && stovetop, sunbed, support\_beam, swivel\_chair,\\
    &&   table, table\_stand, tablecloth, tank, telephone, \\
    &&  telescope, tent, throw\_blanket, tile, toilet, tool, \\
    && towel, toy, trampoline, trashcan, tray, \\
    && treadmill, tv, tv\_stand, umbrella, urinal, \\
    && vase, water\_fountain, water\_heater, water\_tank, \\
    &&  window, window\_curtain, window\_frame,\\
    && window\_shade, window\_shutter, wine\_cabinet, \\
    && wood, workstation, worktop, wreath \\
    \bottomrule
  \end{tabular}
  \label{tab:object}
\end{table}
\subsection{Generalization analysis}

To validate the generalization ability of LongNav-R1, we trained exclusively on HM3D and evaluated on all other benchmarks via direct inference. Although trained on only 6 target types in HM3D, LongNav-R1 generalizes to MP3D and OVON, handling 21 and 371 object categories, respectively. Tab~\ref{tab:object} details these categories, highlighting the extensive diversity in OVON. Fig~\ref{fig:category} reports the performance (Success Rate and SPL) across object categories in HM3D (V1/V2), MP3D, and OVON. We can see that: i) for MP3D, there is no significant performance gap between the 6 trained categories and the unseen categories; and ii) for OVON, the model achieves high performance on unseen categories that are even semantically distant from the training set, such as fireplace, piano, flag, cooker. This demonstrates that LongNav-R1 has learned generalizable navigation capabilities rather than merely memorizing object shortcuts.

\begin{figure*}[!t]
\centering
    \includegraphics[width=0.22\linewidth]{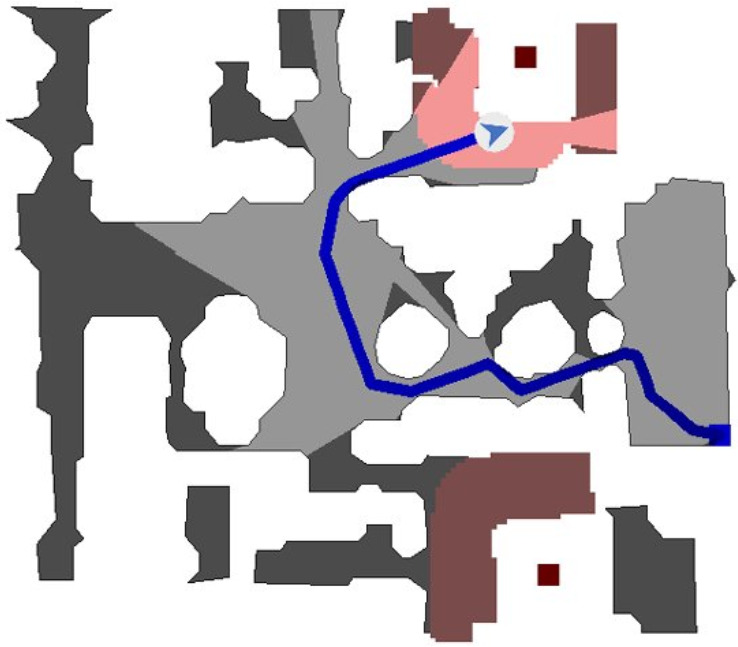}
    \includegraphics[width=0.24\linewidth]{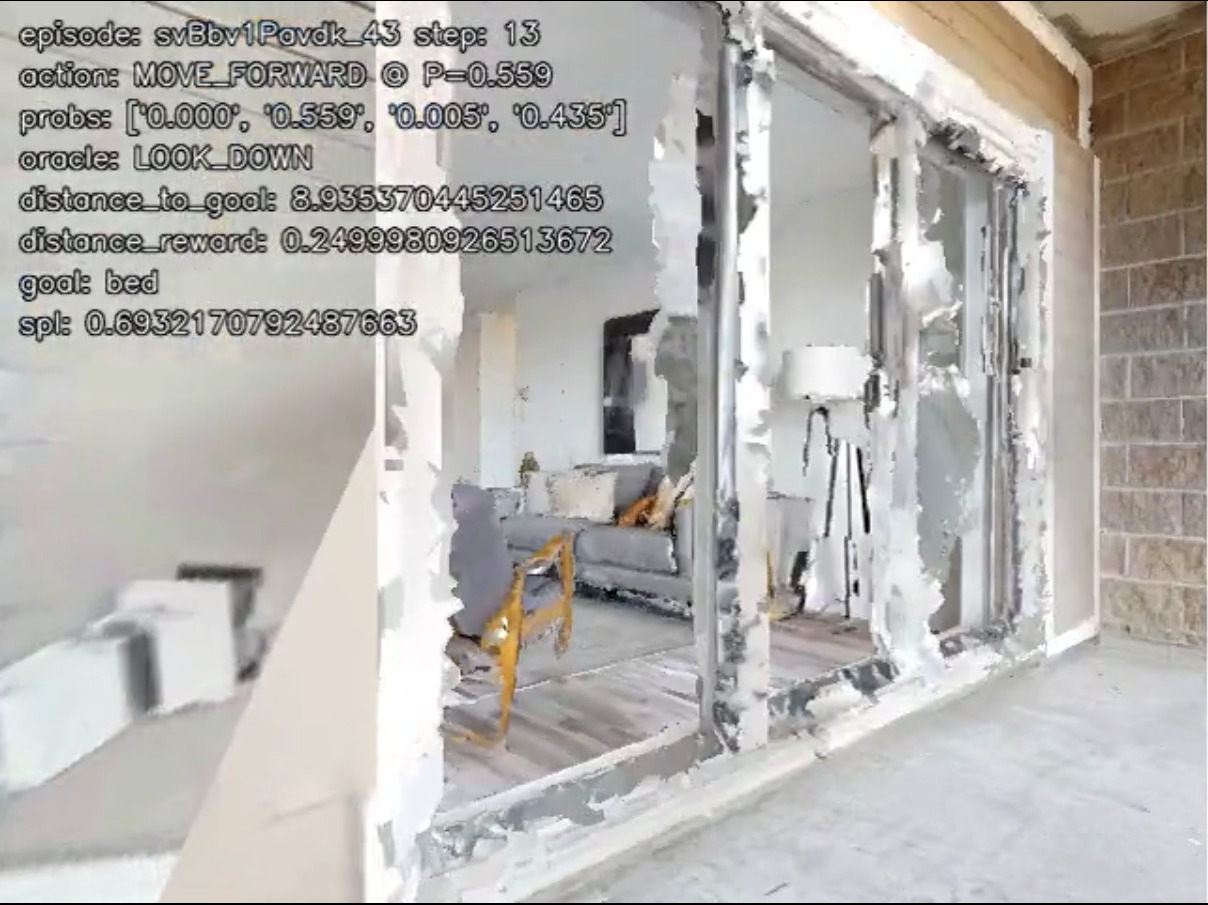}
    \includegraphics[width=0.24\linewidth]{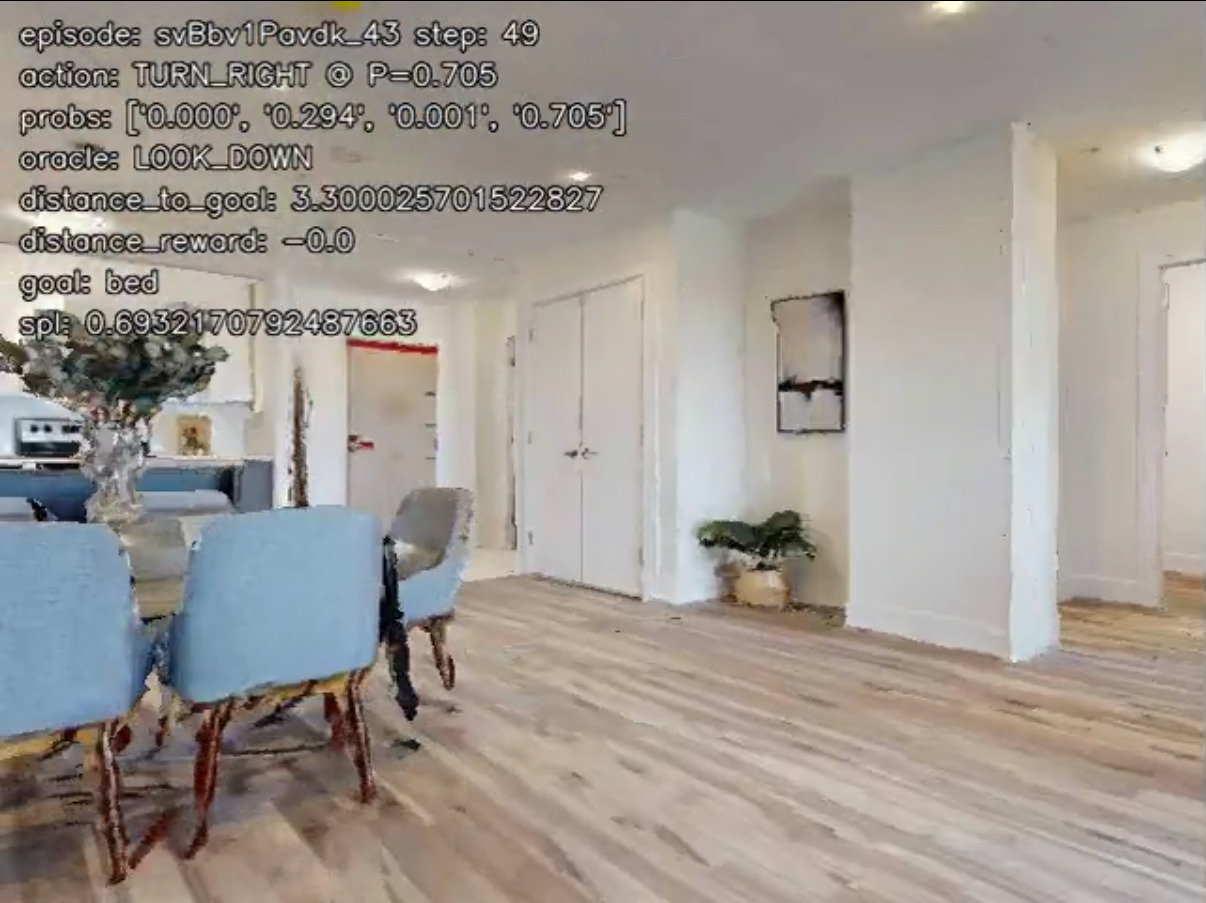}
    \includegraphics[width=0.24\linewidth]{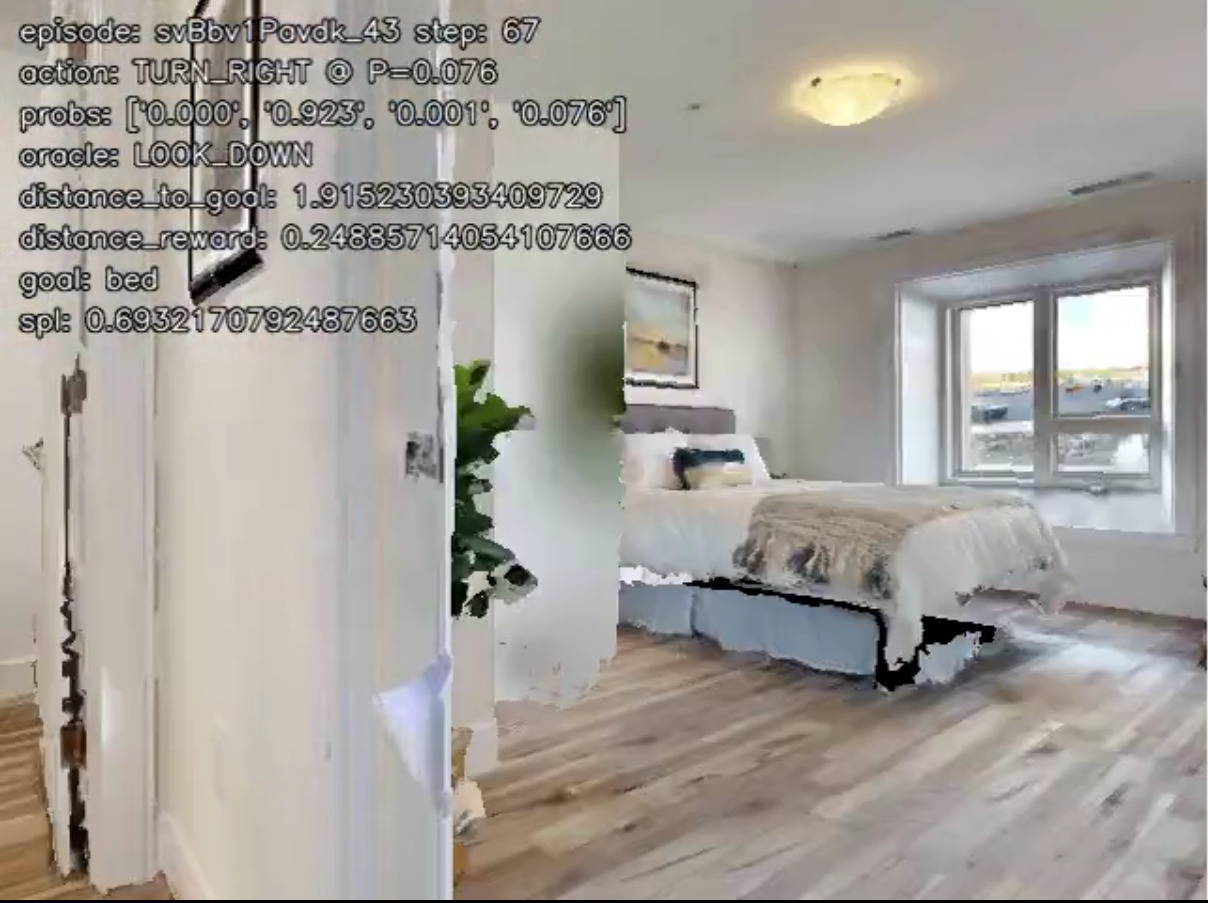}
    \vspace{-1mm}
    \caption{Visualization of navigation episode. The agent successfully navigates across rooms for 81 steps to locate the target bed. It starts on the balcony, travels indoors to the dining room, and discovers the bed in a nearby bedroom.}
    \label{fig:case_study_100a}
    \vspace{-3mm}
\end{figure*}

\begin{figure*}[!t]
\centering
    \includegraphics[width=0.50\linewidth]{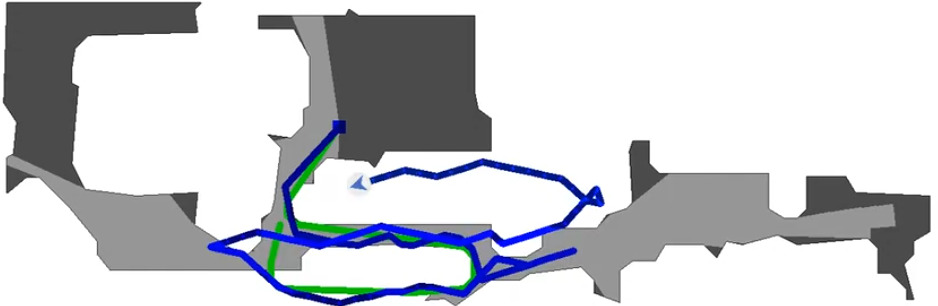}
    \includegraphics[width=0.25\linewidth]{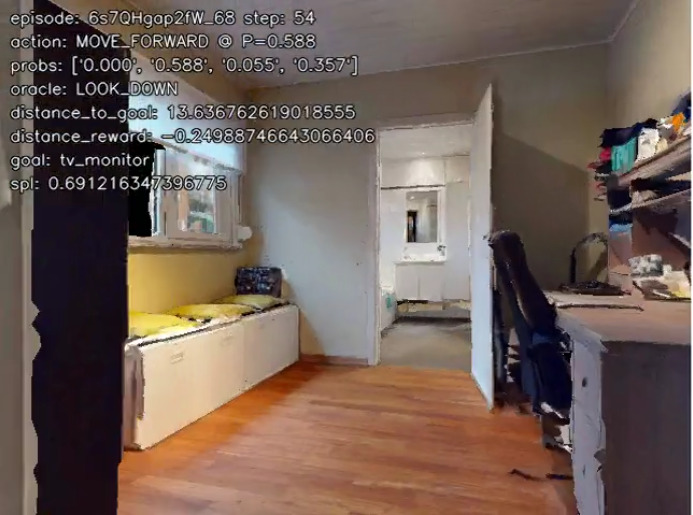}
    \includegraphics[width=0.25\linewidth]{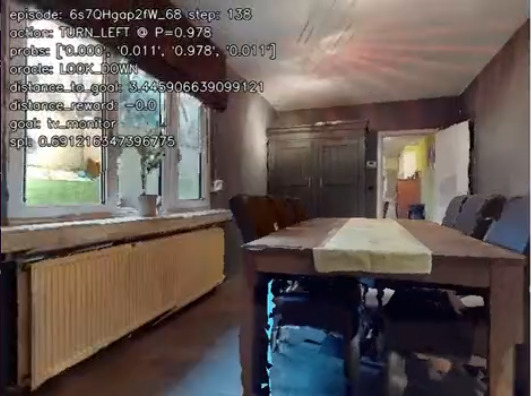}
    \includegraphics[width=0.25\linewidth]{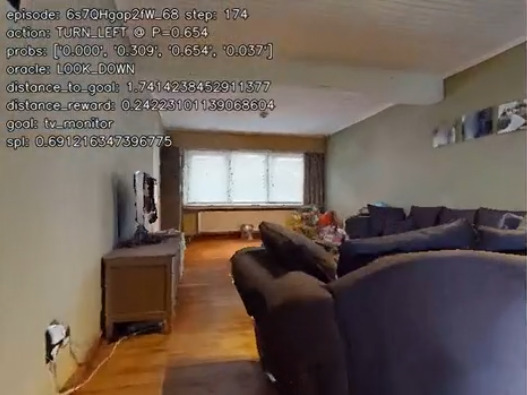}
    \includegraphics[width=0.25\linewidth]{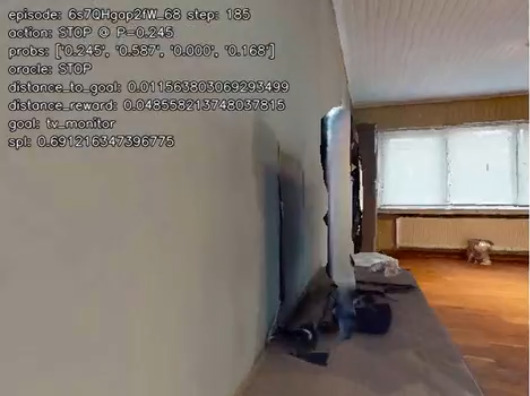}
    \vspace{-1mm}
    \caption{Visualization of a long-horizon navigation episode. The agent successfully navigates across floors for 185 steps to locate the target tv monitor. Starting at the second floor, agent travels across a corridor to check that no tv is present in the study or bathroom. It then travels downstairs, traverses the dining room, and finds the tv in the adjacent living room. }
    \label{fig:case_study_200a}
    \vspace{-3mm}
\end{figure*}

\begin{figure*}[!t]
    \centering
        \includegraphics[width=0.25\linewidth]{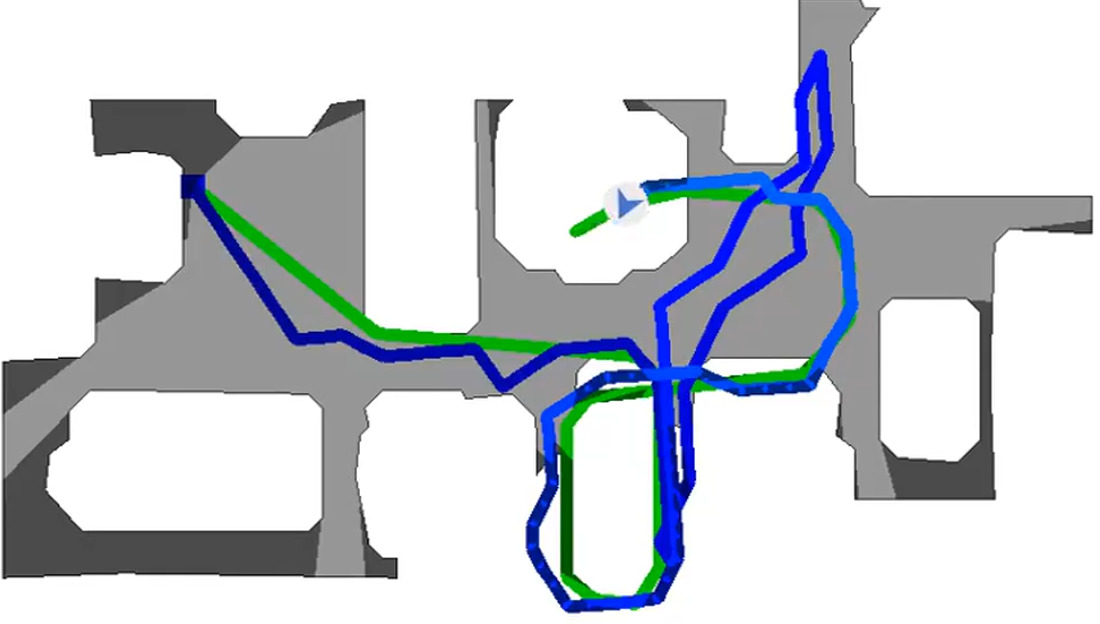}
        \includegraphics[width=0.25\linewidth]{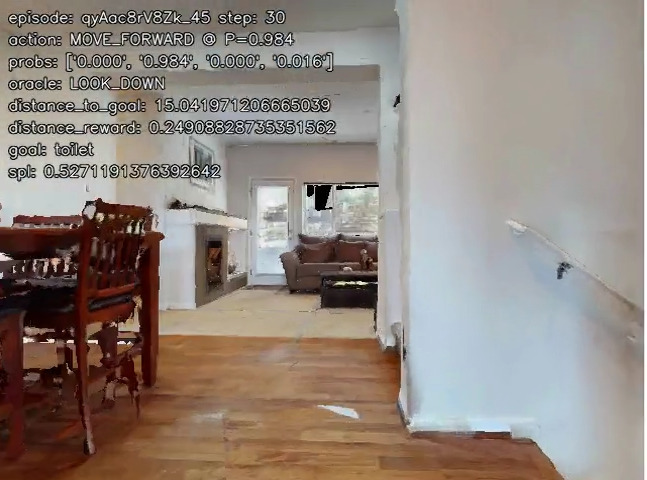}
        \includegraphics[width=0.25\linewidth]{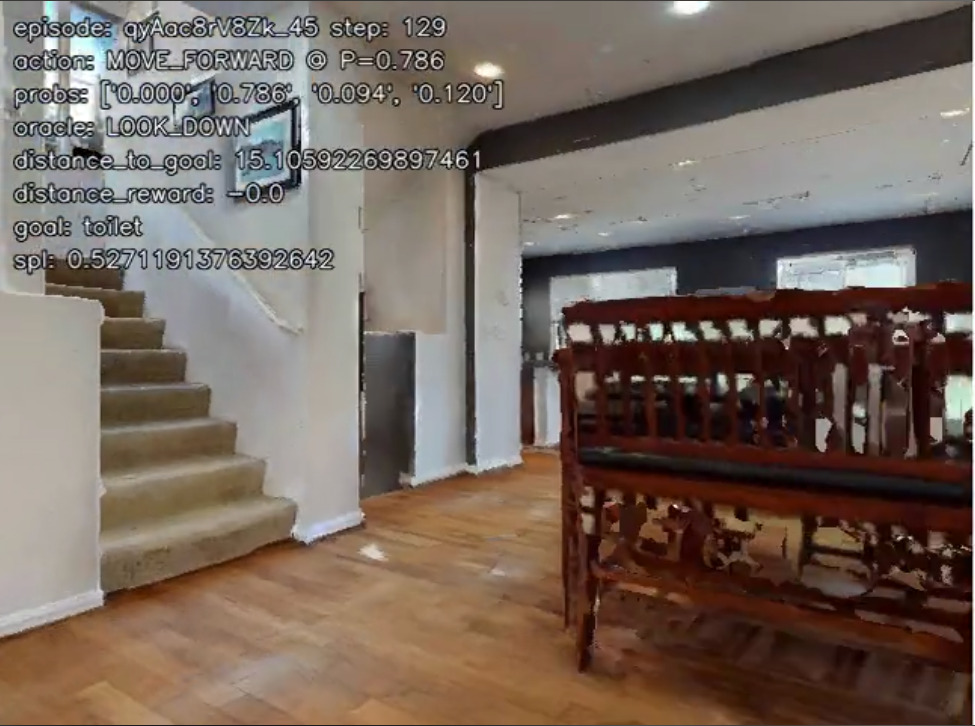}
        \includegraphics[width=0.25\linewidth]{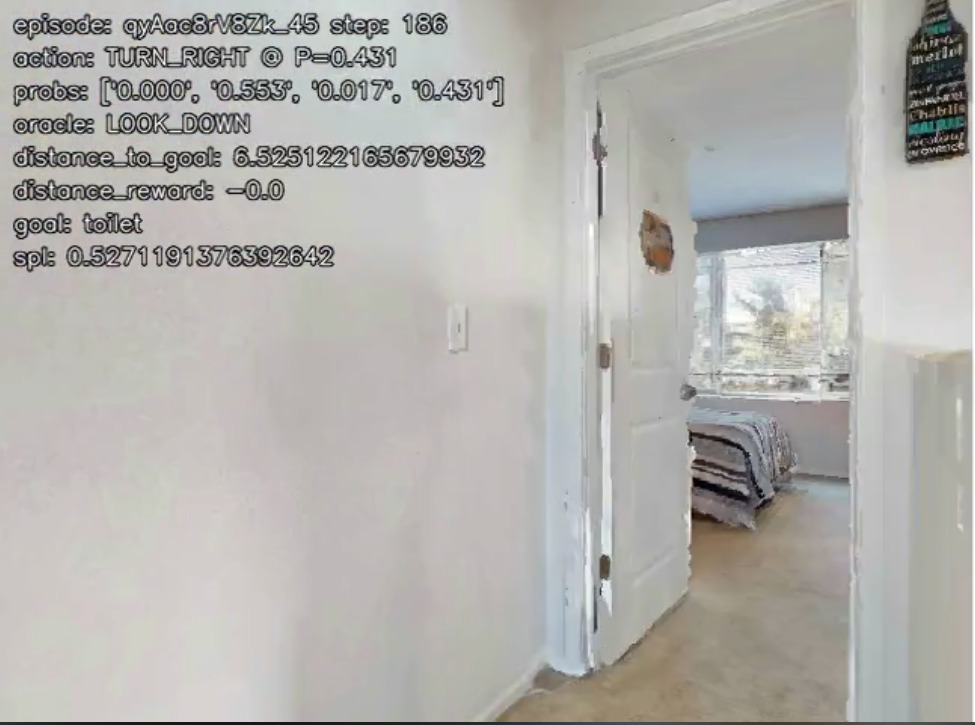}
        \includegraphics[width=0.25\linewidth]{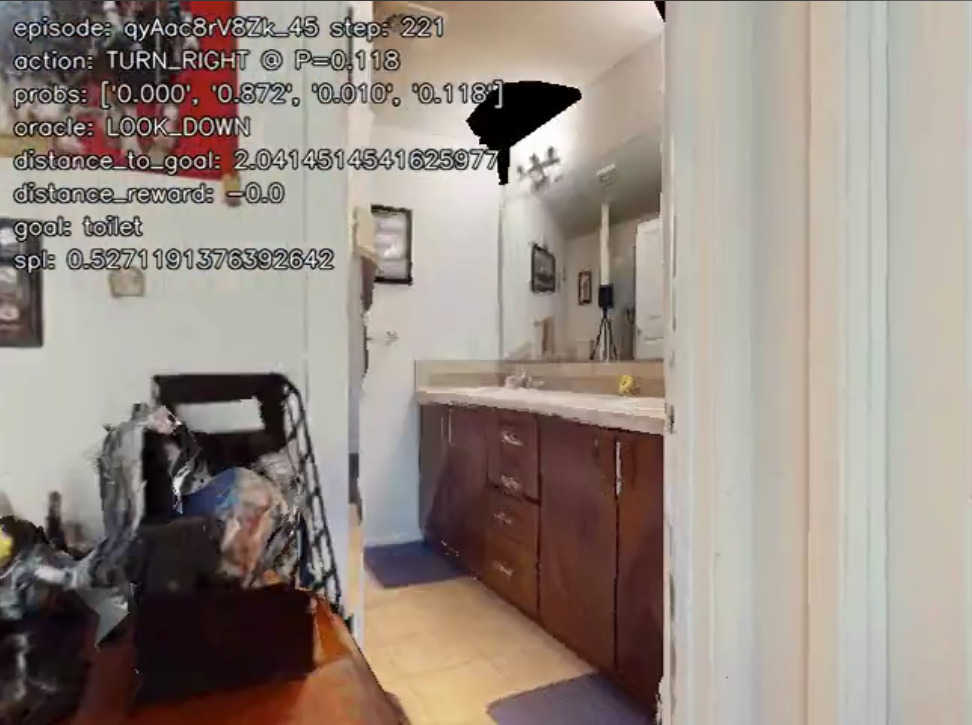}
        \includegraphics[width=0.25\linewidth]{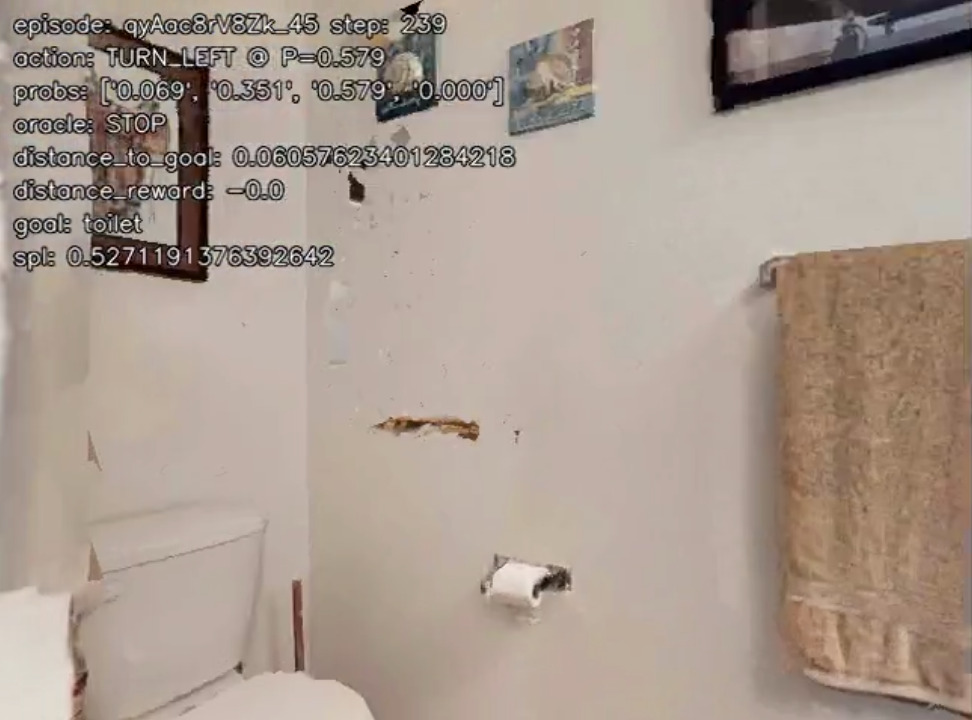}
    \vspace{-1mm}
    \caption{Visualization of a long-horizon navigation episode. The agent successfully navigates across floors for 239 steps to locate the target toilet. The agent explores the first floor thoroughly, then climbs to the second floor. It peeks into a bedroom without entering, and moves on to find the bathroom containing the toilet.}
    \label{fig:case_study_200b}
    \vspace{-3mm}
\end{figure*}

\begin{figure*}[!t]
    \centering
        \includegraphics[width=0.75\linewidth]{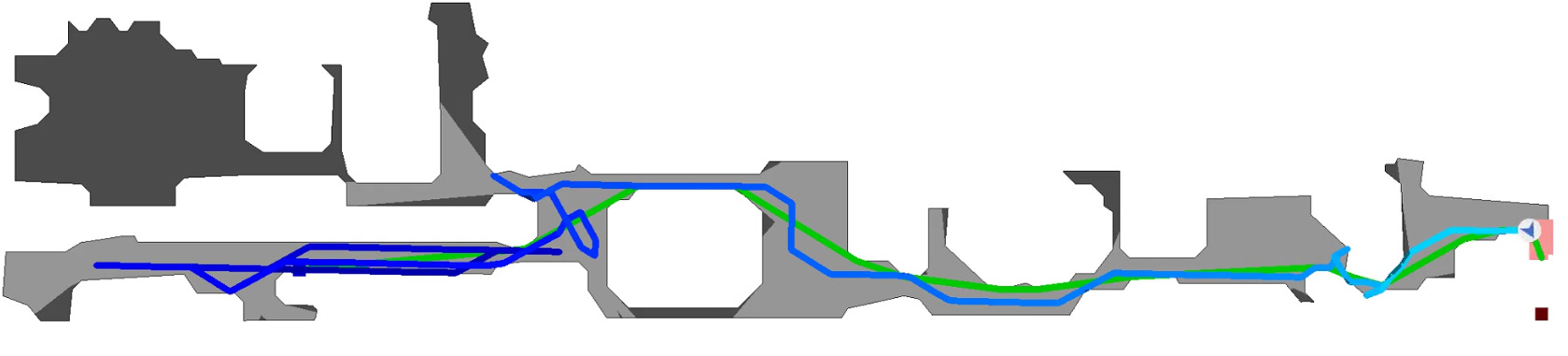}
        \includegraphics[width=0.25\linewidth]{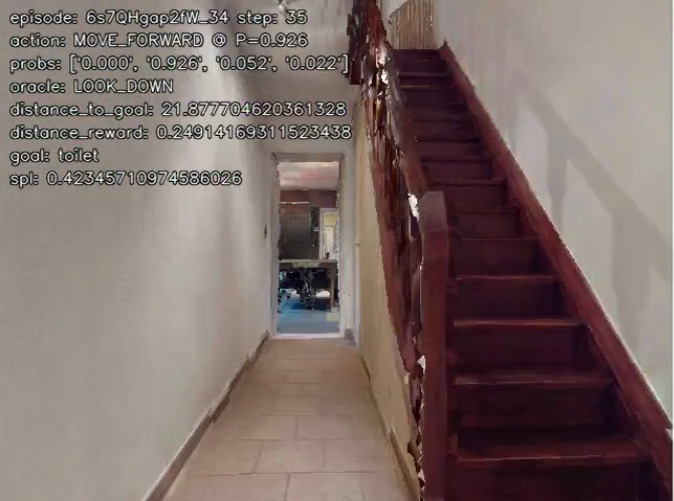}
        \includegraphics[width=0.25\linewidth]{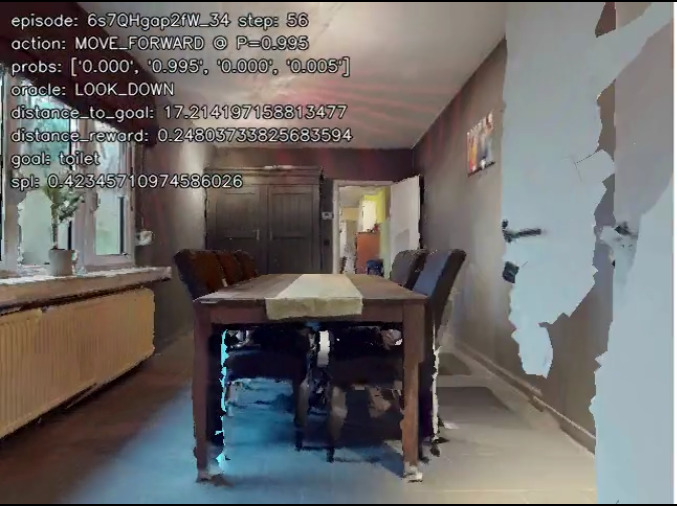}
        \includegraphics[width=0.25\linewidth]{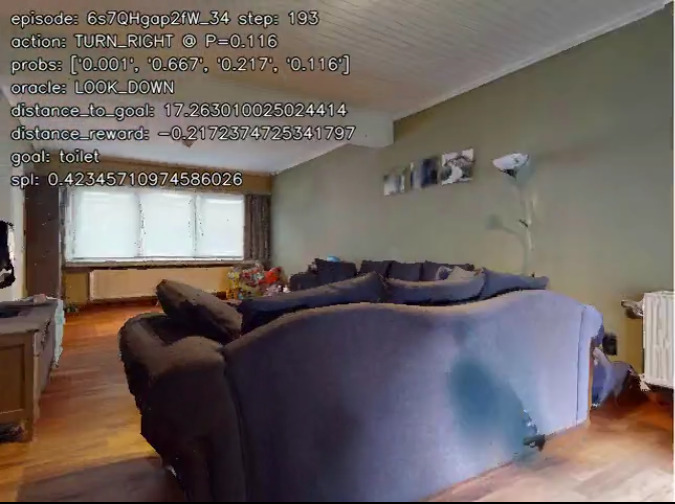}
        \includegraphics[width=0.25\linewidth]{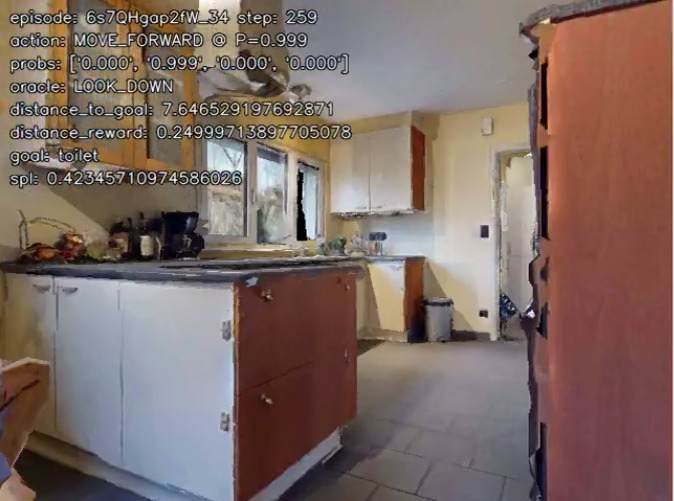}
        \includegraphics[width=0.25\linewidth]{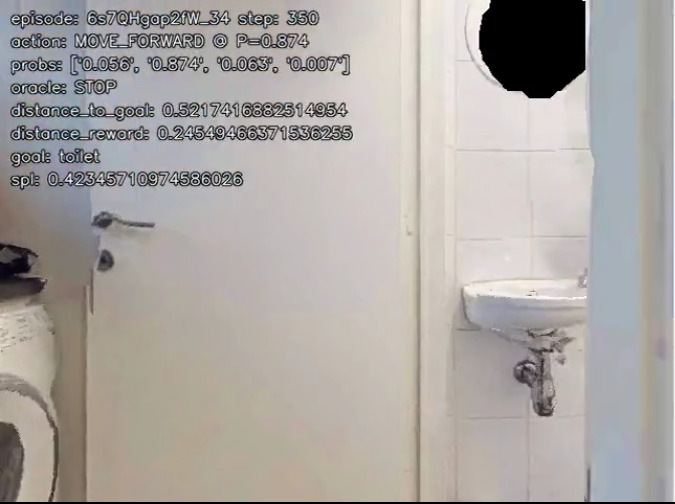}
        \includegraphics[width=0.25\linewidth]{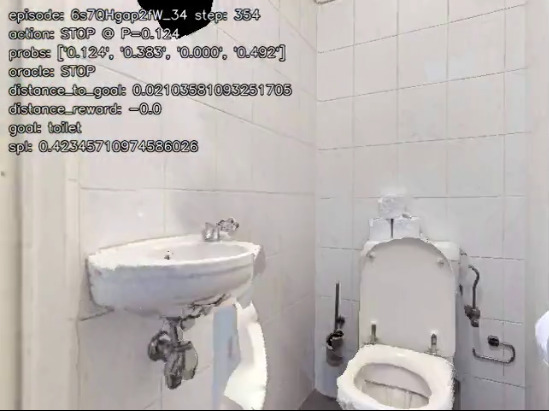}
    \vspace{-1mm}
    \caption{Visualization of a long-horizon navigation episode. The agent successfully explores the oblong first floor for 354 steps to locate the target toilet. The agent begins near the house entrance and proceeds down the entrance corridor. It crosses through the kitchen, turns to peek into the living room without entering, and turns back to continue though the kitchen and into the adjacent laundry room. Finally, it finds the toilet in the bathroom connected to the laundry room.}
    \label{fig:case_study_200c}
    \vspace{-3mm}
\end{figure*}

\subsection{Long-horizon analysis}

Fig.~\ref{fig:case_study_100a},~\ref{fig:case_study_200a},~\ref{fig:case_study_200b} and~\ref{fig:case_study_200c} shows the navigation process with increasing steps, ranging from 81 to 354 steps. LongNav-R1 demonstrates the ability to locate distant targets by sustaining long-horizon exploration and avoiding repetitive trajectories. 

We identify the representative long horizon trajectory examples in ~\ref{fig:case_study_200a},~\ref{fig:case_study_200b} and~\ref{fig:case_study_200c} by filtering for episodes where the oracle horizon (2 times the steps taken by shortest path oracle) is greater than 200 steps. In these challenging episodes where long term consistency is required to reach the goal, LongNav-R1 retains effective active perception strategies such as peeking and panoramic scanning, enabling it to capitalize on promising regions and avoid dead ends.

% Furthermore, LongNav-R1 retains effective active exploration strategies over long trajectories, discovering promising regions and avoiding dead ends via peeking and panoramic scanning behaviors.

% \subsection{Failure case analysis}

% Each show 2 samples, each sample contains 4 sub-figures.

% sub-figure 1: BEV figure including target, GT path, inferred path, sub-figure 2-4: important decision points and final stop image. Show success case. 

% GT issue.

% Repeat behavior.

% Out of time.

% Stop.

% 8 figures in total.

% \subsection{Ablation studies}

% \noindent\textbf{Performance on VLN.}

% \input{table/r2r}

% \noindent\textbf{Effectiveness of online token pruning.}

% \noindent\textbf{Effectiveness of buffer size.}

% \noindent\textbf{Effectiveness of discount factor.}

% \input{content/08-rebuttal}

\end{document}